\documentclass[twoside]{article}

\usepackage[accepted]{aistats2025}
\allowdisplaybreaks

\usepackage[round]{natbib}

\usepackage{amsmath,amssymb,amsthm,dsfont}
\usepackage{graphicx,caption,subcaption}
\usepackage[colorlinks=true,citecolor=blue,linkcolor=blue,urlcolor=blue]{hyperref}

\renewcommand\thesubfigure{\Roman{subfigure}}

\newtheorem{theorem}{Theorem}
\newtheorem{proposition}{Proposition}
\newtheorem{lemma}{Lemma}

\begin{document}

%

%

\twocolumn[

\aistatstitle{The Uniformly Rotated Mondrian Kernel}

\aistatsauthor{Calvin Osborne \And Eliza O'Reilly}

\aistatsaddress{Harvard University  \And Johns Hopkins University} ]

\begin{abstract}

Random feature maps are used to decrease the computational cost of kernel machines in large-scale problems. The Mondrian kernel is one such example of a fast random feature approximation of the Laplace kernel, generated by a computationally efficient hierarchical random partition of the input space known as the Mondrian process. In this work, we study a variation of this random feature map by applying a uniform random rotation to the input space before running the Mondrian process to approximate a kernel that is invariant under rotations. We obtain a closed-form expression for the isotropic kernel that is approximated, as well as a uniform convergence rate of the uniformly rotated Mondrian kernel to this limit. To this end, we utilize techniques from the theory of stationary random tessellations in stochastic geometry and prove a new result on the geometry of the typical cell of the superposition of uniformly rotated Mondrian tessellations. Finally, we test the empirical performance of this random feature map on both synthetic and real-world datasets, demonstrating its improved performance over the Mondrian kernel on a dataset that is debiased from the standard coordinate axes.

\end{abstract}

\section{INTRODUCTION}

Random feature kernel approximations were introduced by \cite{RahimiRecht} to mitigate the high computational cost of kernel methods in large-scale problems. Instead of using a kernel to implicitly lift data using some map $\phi$ into an infinite dimensional feature space, they proposed an explicit embedding of the data using a low-dimensional \emph{random} feature map $z$ such that the inner product in this feature space approximates the kernel evaluation:
\[K(x, x') = \langle \phi(x), \phi(x') \rangle \approx z(x)^Tz(x').\]
\cite{RahimiRecht} proposed two different random feature maps in their work: \emph{Fourier random features}, which are obtained using Bochner's theorem for stationary kernels, and \emph{random binning features}, which are obtained by partitioning the input space with random axis-aligned hyperplanes and generating a feature map indicating the cell of the partition the input is contained in. The former has received much more attention in subsequent literature, see the recent survey by \cite{Liuetal_Survey_2022} and the references therein. However, it has been observed by \cite{BinningWuetal} that random binning features have computational advantages and improved performance over Fourier random features for several tasks. 

Another random feature kernel generated using random partitions of the input space was proposed by \cite{Balog2016} and called the \textit{Mondrian kernel}. This kernel is constructed by using the random binning feature map induced by a random hierarchical partition called the \textit{Mondrian process} \citep{BalogTeh2015} that has many appealing properties such as the Markov property and spatial consistency. This partition produces a feature map approximating the Laplace kernel, the same kernel that is approximated by the random binning feature map \citep{RahimiRecht}. However, the special properties of the process provide an efficient bandwidth learning procedure \citep{Balog2016}, a vital parameter tuning step that requires expensive cross-validation procedures in the case of kernel methods, Fourier random features, and random binning features. In addition, Mondrian forests and other oblique variants have been shown to achieve minimax-optimal convergence rates in regression problems \citep{mourtada2020minimax, Cattaneo2023, OReillyTranMinimax2021}.

However, one drawback of the Mondrian kernel and random binning features kernel is that the expressiveness of these random feature maps is limited since the partitions are restricted to axis-aligned splits. Indeed, as mentioned above, the Mondrian kernel and the random binning features kernel studied in \cite{BinningWuetal} are both known to approximate the Laplace kernel, whereas Fourier random features have the power to approximate \emph{any} stationary, positive definite, symmetric kernel. 
Recent work by \cite{OReilly2022} has shown that a much larger class of kernels can be approximated by using oblique random partitions, increasing the flexibility of this random feature map. In particular, this work studied the random binning feature map generated by stable under iteration (STIT) processes \citep{Nagel2005} in stochastic geometry. These stochastic processes are quite expressive, as they are defined for any dimension $d$ and are indexed by a probability measure on the unit sphere $S^{d-1}$, called the \emph{directional distribution}, that contains $d$ linearly independent unit vectors in its support. For example, it was observed that the translation and rotation invariant exponential kernel could be approximated when this directional distribution is the uniform measure on the unit sphere. This class of STIT processes includes the Mondrian process as a particular case when the directional distribution is the uniform measure on the standard basis vectors.

Learning methods utilizing random partitions with oblique splits have been shown to obtain improved empirical performance over axis-aligned versions for many datasets \citep{blaser2016random, TehRTFs2019}. However, a major hurdle in adopting these approaches is the increased computational costs in generating splits that depend on linear combinations of up to all $d$ dimensions on the input instead of just one dimension in the axis-aligned case. For example, it was shown by \cite{TehRTFs2019} that isotropic STIT processes generate random forests that obtain improved empirical performance over Mondrian forests, but there are significant computational difficulties in generating an isotropic STIT process rather than a Mondrian process. 

\cite{OReilly2022} showed that any STIT process with a discrete directional distribution can be obtained by lifting to a higher dimensional space and running a Mondrian process. However, the induced class of kernels does not include an isotropic one. This work proposes a variant of the Mondrian kernel that approximates a kernel that is invariant under rotation without implementing a complex isotropic STIT process. In particular, we study the random feature map generated by random partitions with oblique splits by applying a uniformly random rotation to the input space and then partitioning the transformed space with a Mondrian process. We call the associated random feature kernel the \emph{uniformly rotated Mondrian kernel}. 

Our theoretical contributions in this work include characterizing the \emph{isotropic} kernel that is approximated by the uniformly rotated Mondrian kernel as the number of random features approaches infinity, as well as a uniform rate of convergence to this limiting kernel. Similar to the results in \cite{OReilly2022}, these findings will depend on machinery from the theory of stationary random tessellations in stochastic geometry \citep{Schneider2008}, as the superposition of uniformly rotated Mondrian tessellations is composed of cells that are more general polytopes than axis-aligned boxes.

Beyond these theoretical contributions, we empirically study the performance of the uniformly rotated Mondrian kernel on synthetic and real-world data sets. To evaluate the performance of this random feature kernel in comparison to Fourier random features, random binning features, and the Mondrian kernel, we first run comparable experiments on the same datasets that were studied in the proposal of the Mondrian kernel in \cite{Balog2016}. These experiments show that the uniformly rotated Mondrian kernel can achieve similar performance to the Mondrian kernel on what we will show to be an \textit{adversarial} \texttt{CPU} dataset without sacrificing computational efficiency. We then evaluate the performance of the uniformly rotated Mondrian kernel and the Mondrian kernel on a non-adversarial dataset, the \textit{Mondrian line}, to show that the uniformly rotated Mondrian kernel can indeed outperform the Mondrian kernel on a dataset that is debiased from a small number of coordinate axes. In particular, these experiments demonstrate that the uniformly rotated Mondrian kernel exhibits the improved empirical performance observed in other random partition kernels with oblique splits \citep{blaser2016random, TehRTFs2019} without an expensive increase in computational demands as with the isotropic STIT process.

\subsection{Related Work}

Methods that utilize oblique splits have shown improved empirical performance across many tasks in machine learning, but general polyhedral cells incur increased computational and theoretical difficulties. Multiple works have proposed generating a partition by randomly rotating the feature space to mitigate this cost and then partitioning with axis-aligned splits. For example, random rotations have been used to increase the diversity of estimators in ensemble methods \citep{blaser2016random} and show improved performance over axis-aligned random forests. Randomly rotated $k{\text -}d$ trees were studied by \cite{vempala:LIPIcs:2012:3847} for nearest neighbor search and were shown to adapt to the intrinsic dimension of the input data. Our work studies this kind of approach when generating random features with a Mondrian process. This allows us to obtain a closed-form expression of the limiting kernel, and convergence guarantees due to the connection with random tessellation models in stochastic geometry.

\section{BACKGROUND}

First, we will detail some background on random tessellations and, in particular, the Mondrian tessellation. 

A \textit{tessellation} $X$ is a locally finite, countable collection of nonempty compact, convex subsets of $\mathbb{R}^d$ such that
\begin{equation*}
    \bigcup_{K \in X} K = \mathbb{R}^d
\end{equation*}
and $\text{int}\,K \cap \text{int}\,K' = \emptyset$ for all distinct pairs $K,K' \in X$. A \textit{random tessellation} $X$ is a point process on the space of nonempty compact subsets of $\mathbb{R}^d$ such that $X$ is almost surely a tessellation. 

Of particular interest to us is the \textit{Mondrian tessellation of lifetime} $\lambda > 0$, a random tessellation constructed on an axis-aligned box $\mathcal{X} \subset \mathbb{R}^d$ by the following \textit{Mondrian process}:

\begin{enumerate}
    \item For each $1 \leq n \leq d$, we sample $t_n \sim \text{Exp}\left(\left|\mathcal{X}_n\right|\right)$, where $\left|\mathcal{X}_n\right|$ is the length of the box in the $n$-th dimension, and let $n_{\min} = \text{argmin}_n(t_n)$.
    \item If $t_{n_{\min}} \geq \lambda$, then the construction is finished.
    \item Otherwise, we sample $a \sim \text{Unif}(\left|\mathcal{X}_{n_{\min}}\right|)$ and split $\mathcal{X}$ into boxes $\mathcal{X}^{<} = \{x \in \mathcal{X} \mid x_{n_{\min}} \leq a\}$ and $\mathcal{X}^{>} = \{x \in \mathcal{X} \mid x_{n_{\min}} \geq a\}$.
    \item We run this same process recursively and independently on $\mathcal{X}^{<}$ and $\mathcal{X}^{>}$ with lifetime $\lambda - t_{n_{\min}} > 0$.
\end{enumerate}

\begin{figure}[b!]
    \centering \includegraphics[width=0.25\columnwidth]{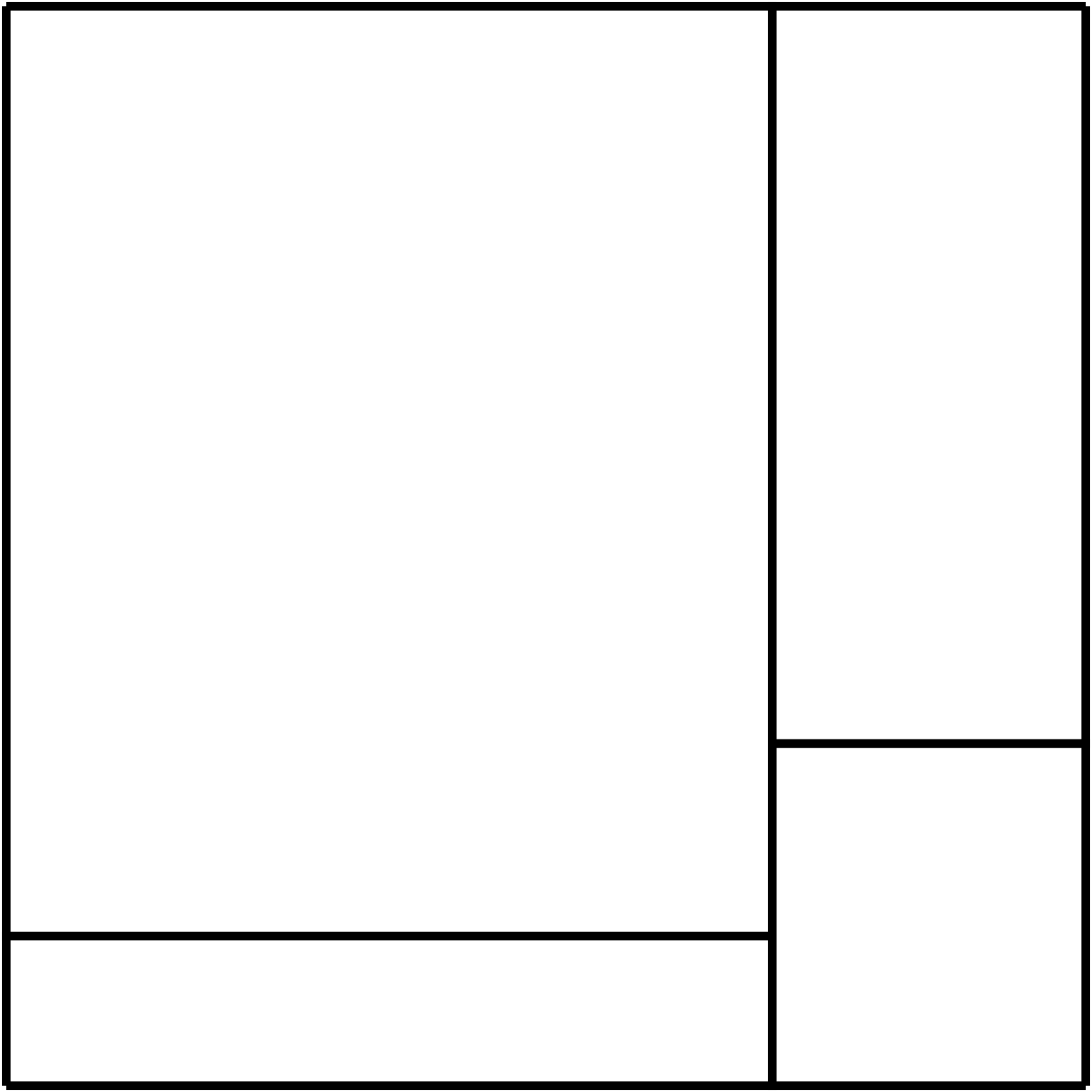}\quad \quad\includegraphics[width=0.25\columnwidth]{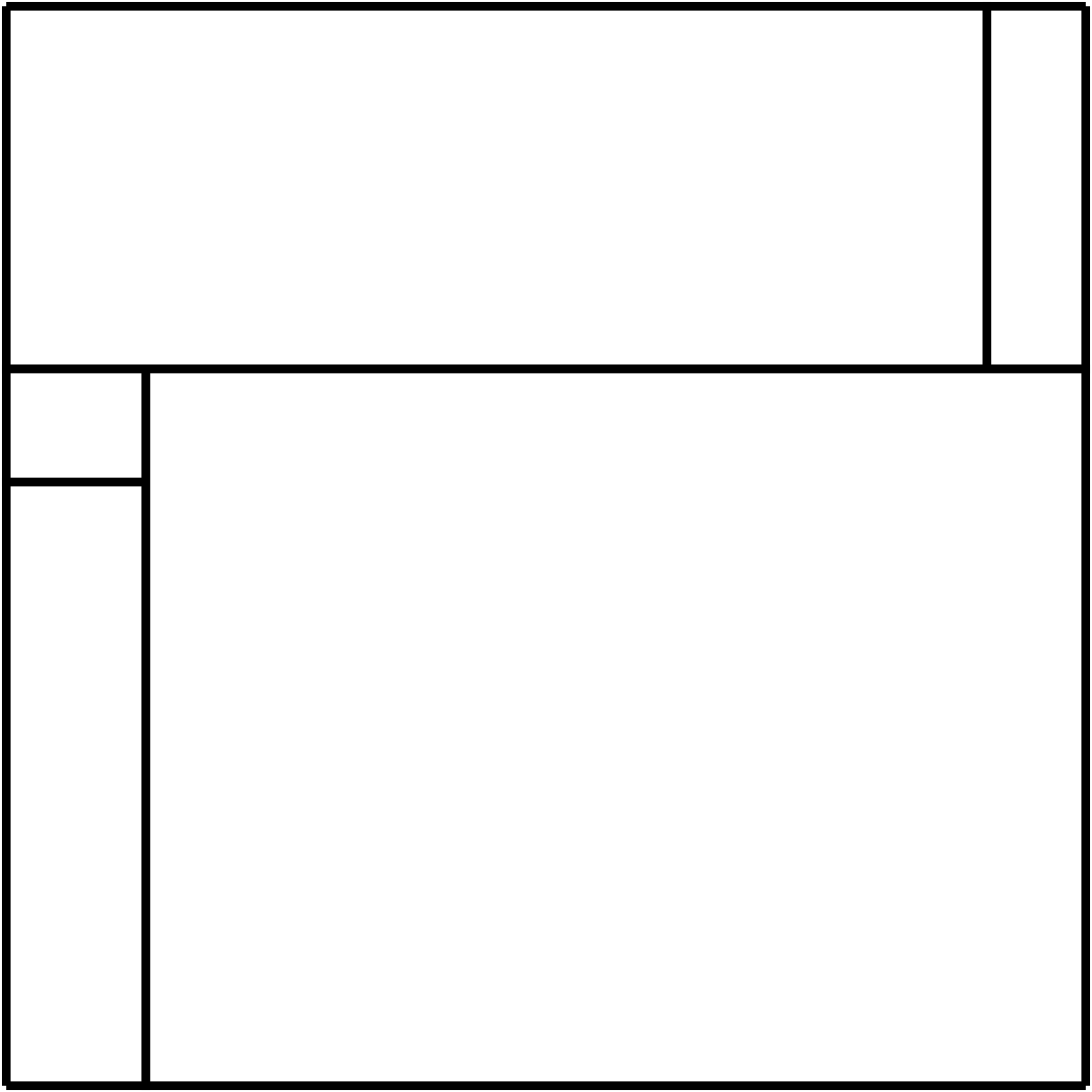}
    \caption{\centering Sample Mondrian Tessellations on $\mathcal{X} = [0, 1]^2$ with $\lambda = 1$.}
    \label{f:Mondrian_sample}
\end{figure}
See Figure \ref{f:Mondrian_sample} for two particular samples of a Mondrian tessellation on $\mathcal{X} = [0, 1]^2$ with lifetime $\lambda = 1$.

\begin{figure}[b!]
    \centering \includegraphics[width=0.25\columnwidth]{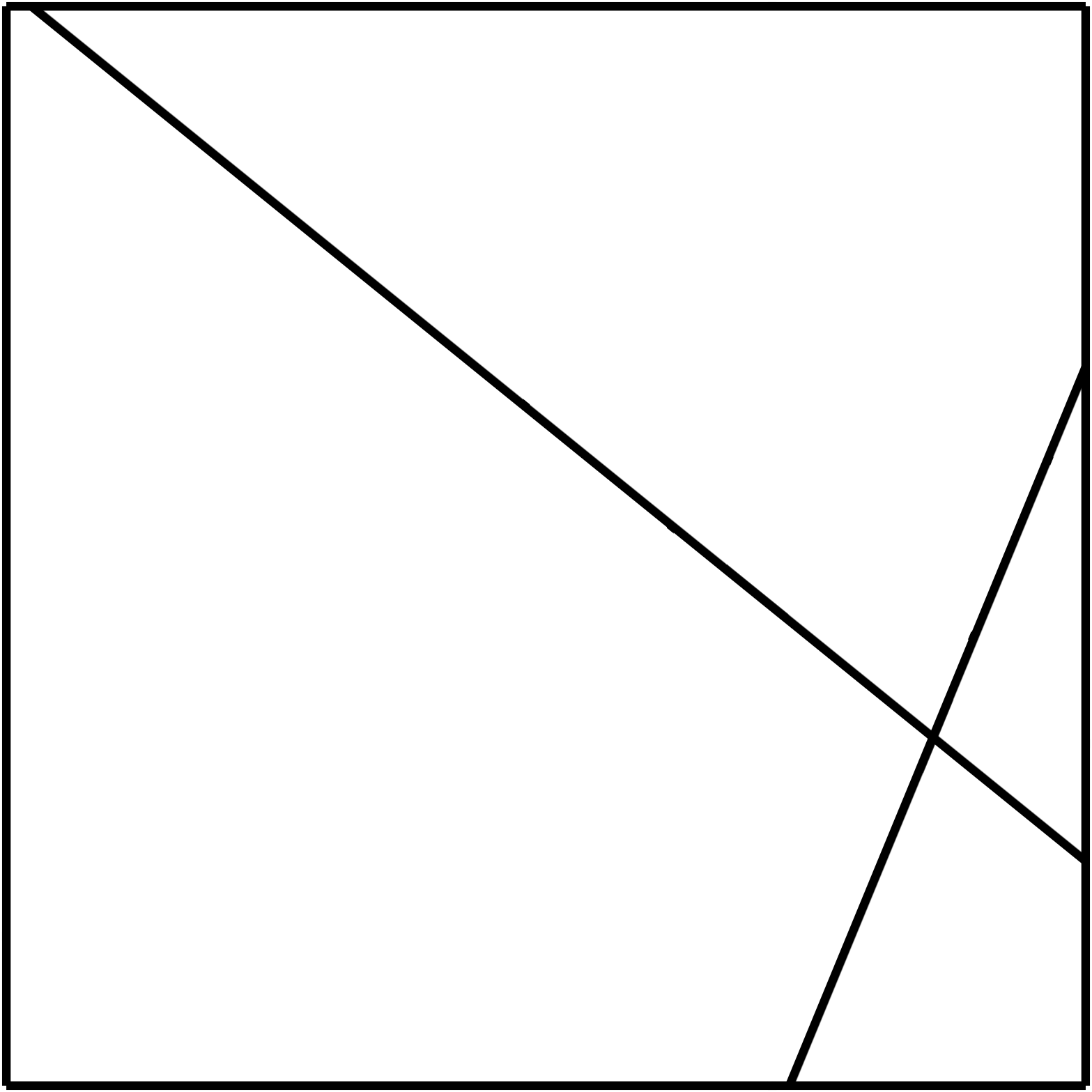}\quad \quad\includegraphics[width=0.25\columnwidth]{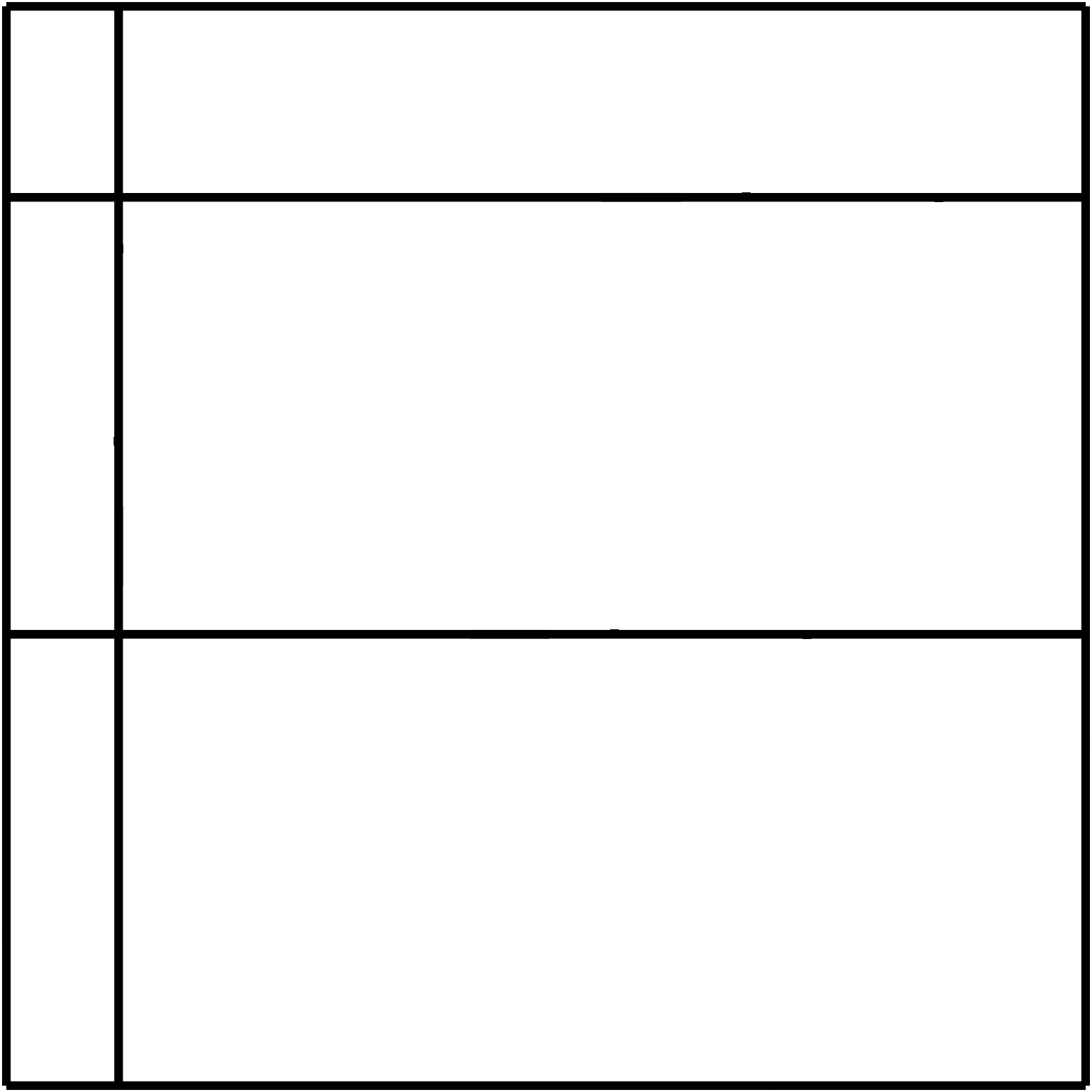}
    \caption{\centering Sample Isotropic Poisson Hyperplane Tessellation and Poisson Manhattan Tessellation on $\mathcal{X} = [0, 1]^2$ with $\lambda = 1$.}
    \label{f:Poisson_sample}
\end{figure}

The second class of random tessellations that will be of interest to us are \textit{stationary Poisson hyperplane tessellations}, a class of random tessellations defined by a stationary Poisson process on the space of hyperplanes in $\mathbb{R}^d$ that constitute the boundaries of their cells. The distribution of these tessellations is determined by a constant intensity $\lambda > 0$ and a directional distribution $\varphi$ on the unit sphere governing the distribution of the normal vectors of the hyperplanes. If $\varphi$ is the uniform distribution on the unit basis vectors in $\mathbb{R}^d$, the random tessellation is called a \textit{Poisson Manhattan tessellation}. See Figure \ref{f:Poisson_sample} for two particular samples of a Poisson hyperplane tessellation on $\mathcal{X} = [0, 1]^2$ with intensity $\lambda = 1$, one with a directional distribution that is uniform on the unit sphere, i.e., an \textit{isotropic Poisson hyperplane tessellation}, and the other a Poisson Manhattan tessellation.

\section{UNIFORMLY ROTATED MONDRIAN KERNEL}

We will now define the \textit{random feature map} $z$ for the uniformly rotated Mondrian process on a dataset $\{x_n\}_{1 \leq n \leq N}$ with $x_n \in \mathbb{R}^d$ as follows.
\begin{enumerate}
    \item First, sample a random rotation $R\sim \text{Unif}(SO_d)$, and map each $x_n \mapsto Rx_n$.
    \item Then, let $\mathcal{X}$ be a bounding box for the rotated dataset, and construct a Mondrian tessellation $X$ on $\mathcal{X}$.  Assign to each cell of $X$ an index $1, 2, \dots$.
    \item Finally, define $z(x)$ to be the one-hot encoding of the index of the cell containing $Rx$, i.e., $z(x) \in \mathbb{R}^C$ where $C$ is the number of cells in $X$, and $z(x)$ has a single coordinate value of one and zero elsewhere.
\end{enumerate}
With this construction of a random feature map $z : \mathbb{R}^d \to \mathbb{R}^C$, we compute the \textit{kernel map} $k : \mathbb{R}^d \times \mathbb{R}^d \to \mathbb{R}$ as
\begin{align*}
    k(x, x') &= z(x)^T z(x') \\[0.5em]
    &= \begin{cases}
        1 & \text{if }Rx, Rx'\text{ are in the same cell of }X, \\
        0 & \text{otherwise}.
    \end{cases}
\end{align*}
Finally, we extend this definition by generating $M$ uniformly rotated Mondrian kernel maps $k^{(1)}, \dots k^{(M)}$ induced by $M$ independently generated uniform rotations $R^{(1)}, \ldots, R^{(M)}$ and Mondrian tessellations $X^{(1)}, \ldots, X^{(M)}$, and we define $k_M : \mathbb{R}^d \times \mathbb{R}^d \to \mathbb{R}$ to be
\begin{equation*}
    k_M(x, x') = \dfrac{1}{M} \sum_{m=1}^M k^{(m)}(x, x').
\end{equation*}
We call $k_M$ the \textit{uniformly rotated Mondrian kernel of order $M$}. By the law of large numbers, the sequence of uniformly rotated Mondrian kernels $k_M$ of order $M$ converges almost surely as $M \to \infty$ to the limiting kernel $k_{\infty}$ given by
\begin{align}\label{e:k_inf}
   k_{\infty}(x, x') &= \mathbb{E}[k(x,x')] \nonumber \\[1em]
   &= \mathbb{P}[Rx,Rx' \text{ in the same cell of } X], 
\end{align}
i.e., given by the probability that the inputs are in the same cell of a uniformly rotated Mondrian tessellation.

\section{MAIN RESULTS}

We will now detail and discuss the two main theoretical results in this paper, with proofs to follow below.

\subsection{Limiting Uniformly Rotated Mondrian Kernel}

Our first result is an expression for the limiting kernel $k_{\infty}$ of the uniformly rotated Mondrian process as is defined in \eqref{e:k_inf}. 

\begin{theorem}\label{thm:limiting_kernel} 
The limiting kernel of the uniformly rotated Mondrian process with lifetime $\lambda > 0$  in $\mathbb{R}^d$ is of the form
\begin{equation}\label{e:limiting_kernel}
    k_\infty(x, x') = \dfrac{1}{\omega_d}\int_{S^{d-1}} e^{-\lambda\left\lVert x - x'\right\rVert_{2}\left\lVert v\right\rVert_{1}}\mathrm{d}v,
\end{equation}
where $\omega_d$ is the surface area of the unit ball in $\mathbb{R}^d$.
\end{theorem}

\begin{figure}[b!]
    \centering \includegraphics[width=0.5\columnwidth]{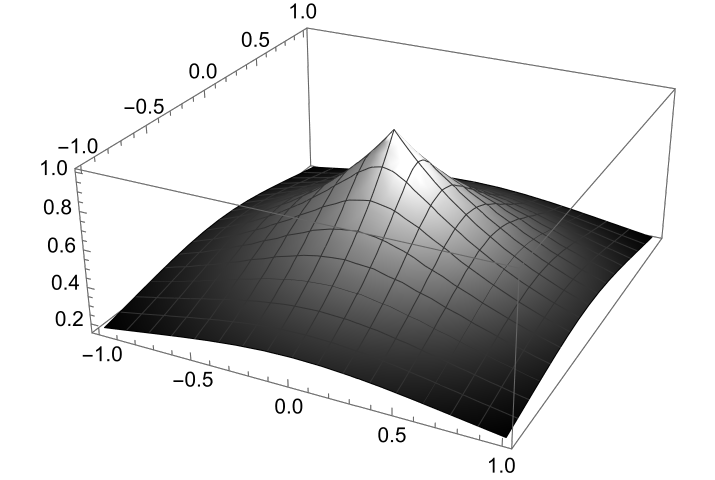}
    \caption{\centering Limiting Kernel for the Uniformly Rotated Mondrian Process on $\mathbb{R}^2$ with lifetime $\lambda = 1$.}
    \label{f:limiting_kernel}
\end{figure}

We note that \eqref{e:limiting_kernel} is indeed isotropic since it depends only on $\lVert x - x'\rVert_{2}$, unlike the limiting Laplace kernel of the standard Mondrian process in \cite{Balog2016} which depends only on $\lVert x - x'\rVert_{1}$. See Figure \ref{f:limiting_kernel} for a plot of this kernel in $\mathbb{R}^2$ with lifetime $\lambda = 1$. As will be detailed in Section \ref{proofs}, this result follows from the computation of an integral over the special orthogonal group $SO_d$ with respect to the normalized Haar measure $\mu$ on $SO_d$.

\subsection{Uniform Convergence of Uniformly Rotated Mondrian Kernel}

Our second result characterizes the rate of convergence of the sequence $k_M$ to $k_\infty$ as $M \to \infty$.

\begin{theorem}\label{thm:rate}
For any bounded domain $\mathcal{X} \subset \mathbb{R}^d$ and small enough $\delta > 0$, we have that
\begin{align}\label{e:rate}
    &\mathbb{P}\left[\sup_{x, x' \in \mathcal{X}} \left|k_M(x, x') - k_\infty(x, x')\right| > \delta\right] \nonumber \\
    &\quad\quad\quad\quad\in \mathcal{O}\left(M^{d + d/(2d + 1)}e^{- M\delta^2/(4d + 2)}\right),
\end{align}
where $k_M : \mathbb{R}^d \times \mathbb{R}^d \to \mathbb{R}$ is the uniformly rotated Mondrian kernel of order $M$.
\end{theorem}

Similar to Proposition 2 in \cite{Balog2016}, this second result guarantees the exponential uniform convergence of $k_M$ to $k_\infty$. Furthermore, we note that this rate of convergence is identical to that of the class of random feature kernels generated by STIT tessellations from Theorem 4.3 in \cite{OReilly2022}.

\section{CELLS OF UNIFORMLY ROTATED MONDRIAN TESSELLATIONS}

To prove Theorem \ref{thm:rate}, we will need to study the geometry of the cells of the random tessellation of the input space generated by the rotations and Mondrian partitions that constitute the uniformly rotated Mondrian kernel. 
In particular, we will study the geometry of the \textit{typical cell}. The typical cell $Z$ of a stationary random tessellation is a random nonempty compact, convex polytope with distribution as in (4.8) of \cite{Schneider2008} given by choosing a cell uniformly on a compact, convex window $rW \subset \mathbb{R}^d$ and letting $r \to \infty$. In this section, we will state three results concerning the volume, inradius, and circumradius of the typical cell. Then, to  prove Theorem \ref{thm:rate}, we will make frequent use of an application of Campbell's theorem as follows.

\begin{theorem} \textup{\citep[Equation 4.3]{Schneider2008}} \label{thm:campbell}
For any nonnegative measurable function $f$ on the space of nonempty compact sets in $\mathbb{R}^d$, and for any stationary random tessellation $X$, we have that
\begin{equation*}
    \mathbb{E}\left[\sum_{C \in \text{cells}(X)}f(C)\right] = \dfrac{1}{\mathbb{E}[V(Z)]} \cdot \mathbb{E}\left[\int_{\mathbb{R}^d} f(Z + y)\mathrm{d}y\right],
\end{equation*}
where $Z$ is the typical cell of the $X$.
\end{theorem}

Conditioned on the uniform rotation $R$, the random tessellation of the input space generated by a uniformly rotated Mondrian kernel of order one is what we will call a \emph{rotated Mondrian tessellation}, that is, a STIT tessellation with directional distribution 
\begin{equation}\label{e:rotated_phi}
    \varphi = \dfrac{1}{d}\sum_{i=1}^d \delta_{u_{i}},
\end{equation}
where $\{u_1, \dots, u_d\}$ is an orthonormal set in $\mathbb{R}^d$ such that $u_i = Re_i$. We will then make use of Theorem 1 from \cite{Schreiber2013} which gives us that the typical cell of a rotated Mondrian tessellation is the same in distribution as the typical cell of a Poisson Manhattan tessellation with directional distribution \eqref{e:rotated_phi} and the same lifetime or intensity parameter $\lambda$. We will call the latter a \emph{rotated Poisson Manhattan tessellation}.

To now understand the asymptotics of the uniformly rotated Mondrian kernel of order $M$, we will need to study the \textit{superposition} of independent uniformly rotated Mondrian tessellations. For any two tessellations $X,X'$, their superposition is the tessellation composed of the pairwise intersection of their compact, convex cells. In the following, we will make use of a result from Page 158 of \cite{Schneider2008} that states that the superposition of independent stationary Poisson processes is itself a stationary Poisson process, with an intensity measure given by the sum of the two component intensity measures. In summary, we will study the typical cell of the superposition of independent uniformly rotated Mondrian tessellations by considering the associated superposition of independent uniformly rotated Poisson Manhattan tessellations, which itself is a stationary hyperplane tessellation generated by a doubly stochastic Poisson hyperplane process. 

With this background in mind, the results that we will establish to prove Theorem \ref{thm:rate} are stated below.

\begin{lemma}\label{thm:cells} 
Let $X$ be the superposition of $M$ independent rotated Poisson Manhattan tessellations $X_1, \dots, X_M$ of intensity $\lambda > 0$ in $\mathbb{R}^d$, and let $Z$ be the typical cell of $X$. First, it holds that
\begin{equation}\label{e:vZ_bnds}
    \dfrac{1}{\kappa_d} \left(\dfrac{2 \sqrt{d}}{\lambda M}\right)^d \leq \mathbb{E}\left[V(Z)\right] \leq \dfrac{1}{\kappa_d} \left(\dfrac{2 d}{M\lambda }\right)^d,
\end{equation}
where $V(C)$ is the volume of a compact, convex  body $C \subset \mathbb{R}^d$ and $\kappa_d$ is the volume of the unit ball in $\mathbb{R}^d$. Second, it holds that
\begin{equation}\label{e:rZ}
    \mathbb{E}[r(Z)] = \dfrac{1}{2M\lambda},
\end{equation}
where $r(C)$ is the inradius of a compact, convex  body $C \subset \mathbb{R}^d$. Finally, it holds that
\begin{equation}\label{e:RZ_bnd}
    \mathbb{P}[R(Z) \geq a] \leq e^{-2 M \lambda a/d} \left(\sum_{n=0}^{d-1}\dfrac{1}{n!} \left[\dfrac{2 \lambda a}{d}\right]^n\right)^M,
\end{equation}
where $R(C)$ is the circumradius of a compact, convex  body $C \subset \mathbb{R}^d$.
\end{lemma}


We will briefly remark here on the methods used to prove Lemma \ref{thm:cells} and will provide the full details in the Appendix. 

Equation \eqref{e:vZ_bnds} makes use of an object known as the \textit{associated zonoid} $\Pi_X$ of a Poisson hyperplane tessellation $X$. The associated zonoid $\Pi_X$ is an explicit compact, convex body that is associated with the general shape of the cells in $X$. In particular, we rely on Theorem 10.3.3 in \cite{Schneider2008} that states that $\mathbb{E}[V(Z)] = V(\Pi_X)^{-1}$.

Equation \eqref{e:rZ} follows from the more general Theorem 10.4.8 in \cite{Schneider2008} that classifies the exact distribution of the inradius of the typical cell of a general Poisson hyperplane tessellation. These results can be applied to our class of superpositions of rotated Mondrian tessellations by Theorem 1 of \cite{Schreiber2013} as was discussed above.

Equation \eqref{e:RZ_bnd}, to the best of our knowledge, makes use of a new technique to provide an explicit bound on the cumulative distribution function of the circumradius of the typical cell of the superposition of rotated Poisson hyperplane tessellations. One comparable result from \cite{Hug2007}, applied by \cite{OReilly2022} to STIT tessellations, gives a bound of the form
\begin{equation*}
    \mathbb{P}[R(Z) \geq a] \leq ce^{- \tau a h_{\text{min}}(\Pi_X)}
\end{equation*}
for constants $c,\tau > 0$, where $h_{\min}(\Pi_X)$ is the minimum of the support function of the associated zonoid $\Pi_X$ of the STIT tessellation $X$. However, since we do not know the dependence of the constant $c > 0$ on the number $M$ of superpositions and the directional distribution $\varphi_n$ of each rotated Mondrian tessellation, this result is not suitable for understanding the limiting behavior of the uniformly rotated Mondrian kernel as $M \to \infty$. As detailed in Proposition \ref{thm:lifting}, our technique of lifting the tessellation $X$ in $\mathbb{R}^d$ to the intersection of a higher-dimensional tessellation $\widetilde{X}$ in $\mathbb{R}^{Md}$ with a $d$-dimensional linear subspace $U$ allows us to compute explicit constants and obtain \eqref{e:RZ_bnd}.

\section{PROOFS OF MAIN RESULTS\label{proofs}}

We will now discuss the proofs of Theorem \ref{thm:limiting_kernel} and Theorem \ref{thm:rate}, with further necessary details to be provided in the Appendix. 

\noindent \textit{Proof of Theorem \ref{thm:limiting_kernel}.} We first compute the limiting kernel of a Mondrian process under a fixed rotation $R \in SO_d$, and we then compute the limiting kernel of the uniformly rotated Mondrian process by conditioning on the rotation $R \in SO_d$. 

For a fixed rotation $R \in SO_d$, letting $k_\infty$ be the rotated Mondrian limiting kernel and $\hat{k}_\infty$ be the Mondrian limiting kernel, we have from Proposition 1 in \cite{Balog2016} that
\begin{align*}
    k_\infty(x, x') &= \hat{k}_\infty(Rx, Rx') = e^{-\lambda \left|\left|R(x - x')\right|\right|_{1}}.
\end{align*}
Letting $\mu$ be the normalized Haar measure on $SO_d$, we compute that
\begin{align*}
    k_\infty(x, x') &= \mathbb{P}[k(x,x') = 1] \\[1em]
    &= \int_{SO_d} \mathbb{P}[\hat{k}(Rx, Rx') = 1] \mathrm{d}\mu(R) \\[1em]
    &= \int_{SO_d} \hat{k}_\infty(Rx, Rx')\mathrm{d}\mu(R) \\[1em]
    &= \int_{SO_d} e^{-\lambda \left|\left| R(x - x')\right|\right|_{1}} \mathrm{d}\mu(R)
\end{align*}
from the law of total probability. Finally, from the translation invariance of the Haar measure $\mu$, we have from right multiplication by the rotation $S \in SO_d$ that points $x - x'$ in the direction of the axis-aligned unit vector $e_1 \in \mathbb{R}^d$ that
\begin{align*}
    k_\infty(x, x') &= \int_{SO_d} e^{-\lambda \left|\left| RS(x - x')\right|\right|_{1}} \mathrm{d}\mu(R) \\[1em]
    &= \int_{SO_d} e^{-\lambda \left|\left|x - x'\right|\right|_{2}\left|\left|Re_1\right|\right|_{1}}\mathrm{d}\mu(R) \\[1em]
    &= \dfrac{1}{\omega_d}\int_{S^{d-1}} e^{-\lambda \left|\left|x - x'\right|\right|_{2}\left|\left|v\right|\right|_{1}} \mathrm{d}v
\end{align*}
as is desired, completing our computation of the limiting kernel of the uniformly rotated Mondrian. \hfill $\square$

We will now provide an overview of the proof of Theorem \ref{thm:rate}.

\textit{Outline of Proof of Theorem \ref{thm:rate}.} We will first let $B_r$ be a ball of radius $r$ in $\mathbb{R}^d$ such that $\mathcal{X} \subseteq B_r$. We will then let $\mathcal{U}$ be an $\varepsilon$-grid covering of a concentric $B_{R + \varepsilon}$ for some $R > r$ and $\varepsilon > 0$ to be specified later. Letting $X$ be the superposition of $M$ independently rotated Mondrian tessellations, we will consider the following four events:
\begin{enumerate}
    \item $A_1 = \{\text{every cell of }X\text{ with a circumcenter}$
    
    \quad\quad\quad\quad$\text{ outside of }B_R\text{ is disjoint from }B_r\}$,
    \item $A_2 = \{\text{every cell of }X\text{ with a circumcenter in }$
    
    \quad\quad\quad\quad$\text{ }B_R\text{ contains a point of }\mathcal{U}\}$,
    \item $A_3 = \{\text{every cell of }X\text{ with a circumcenter}\text{ in}$
    
    \quad\quad\quad\quad$B_R\text{ has a diameter less than or equal}$
    
    \quad\quad\quad\quad$\text{ to }\delta/4\lambda \sqrt{d}\}$,
    \item $A_4 = \{\text{the }\delta/2\text{-approximation holds on }\mathcal{U}\}$,
\end{enumerate}
where the \textit{$\delta$-approximation} on an $\varepsilon$-grid covering $\mathcal{U}$ is that 
\begin{equation*}
    \left|k_M(u, u') - k_\infty(u, u')\right| \leq \delta
\end{equation*}
for all $u,u' \in \mathcal{U}$. It holds (see the Appendix) that $A_1 \cap A_2 \cap A_3 \cap A_4$ implies that the $\delta$-approximation holds on $\mathcal{X}$, and so we have that
\begin{align*}
    &\mathbb{P}\left[\sup_{x, x' \in \mathcal{X}} \left|k_M(x, x') - k_\infty(x, x')\right| > \delta\right] \\[1em]
    &\quad\quad\quad\quad\quad\leq \mathbb{P}(A_1^c \cup A_2^c \cup A_3^c \cup A_4^c) \\[1em]
    &\quad\quad\quad\quad\quad\leq \mathbb{P}(A_1^c) + \mathbb{P}(A_2^c) + \mathbb{P}(A_3^c) + \mathbb{P}(A_4^c)
\end{align*}
from a union bound. It remains to bound the probability of the four events $A_1^c, A_2^c, A_3^c, A_4^c$. For $A_1^c$, letting $c(C)$ be the circumcenter of a nonempty compact, convex set, we note that
\begin{align*}
    \mathbb{P}(A_1^c) &= \mathbb{P}\left[\bigcup_{\text{cells }C \in X}\left(c(C) \not \in B_r\right) \cap (C \cap B_r \neq \emptyset)\right] \\[1em]
    &\leq \mathbb{E}\left[\sum_{\text{cells }C \in X}\mathds{1}[c(C) \not \in B_r]\mathds{1}[C \cap B_r \neq \emptyset]\right].
\end{align*}
By Campbell's theorem and Lemma \ref{thm:cells}, we have that
\begin{align*}
    \mathbb{P}(A_1^c) &\leq \left(\dfrac{1}{\mathbb{E}[V(Z)]}\right) \int_{\left|\left|y\right|\right|_{2} \geq R} \mathbb{P}[(Z + y) \cap B_r \neq \emptyset]\mathrm{d}y \\[1em]
    &\leq \left(\dfrac{1}{\mathbb{E}[V(Z)]}\right) \int_{\left|\left|y\right|\right|_{2} \geq R} \mathbb{P}[r(Z)\geq \left|\left|y\right|\right|_{2} - r]\mathrm{d}y \\[1em]
    &\leq \left(\dfrac{1}{\mathbb{E}[V(Z)]}\right) \int_{\left|\left|y\right|\right|_{2} \geq R} e^{-2M \lambda (y - r)}\mathrm{d}y \\[1em]
    &= \left(\dfrac{\kappa_d^2 d}{2 \lambda M}\right)\left(\dfrac{\lambda M}{2\sqrt{d}}\right)^de^{-2 \lambda M (R -r)},
\end{align*}
where $Z$ is the typical cell of $X$ and $\kappa_d$ is the volume of the unit ball in $\mathbb{R}^d$. Deriving similar bounds for the other events, we find that
\begin{align*}
    \mathbb{P}&\left[\sup_{x, x' \in \mathcal{X}}\left|k_M(x, x') - k_\infty(x, x')\right| > \delta\right] \\[1em]
    < &\left(\dfrac{\kappa_d^2 d}{2 \lambda M}\right)\left(\dfrac{\lambda M}{2 \sqrt{d}}\right)^d e^{-2 \lambda M (R - r)} \\[1em]
    &+ \kappa_d^2\left(\dfrac{\lambda R M}{2\sqrt{d}}\right)^d M \lambda \varepsilon \sqrt{d} \\[1em]
    &+ \kappa_d^2\left(\dfrac{\lambda RM}{2\sqrt{d}}\right)^d \left(\sum_{n=0}^{d-1} \dfrac{1}{n!}\left[ \dfrac{\delta}{4 d^{3/2}} \right]^n\right)^Me^{-M \delta/4d^{3/2}} \\[1em]
    &+ 2 \varepsilon^{-2d} \left(\sum_{n=0}^{2d}\binom{2d}{n}(4R)^n \right)e^{-M\delta^2/2}.
\end{align*}
This expression, when minimized with respect to $\varepsilon$, gives us the desired result for small enough $\delta > 0$. \hfill $\square$

A detailed proof of Theorem \ref{thm:rate} is included in the Appendix.

\section{EXPERIMENTS}

\begin{figure*}[t!]
\renewcommand\thesubfigure{\arabic{subfigure}}
  \begin{subfigure}[b]{0.49\textwidth}
         \centering
         \includegraphics[width=\textwidth]{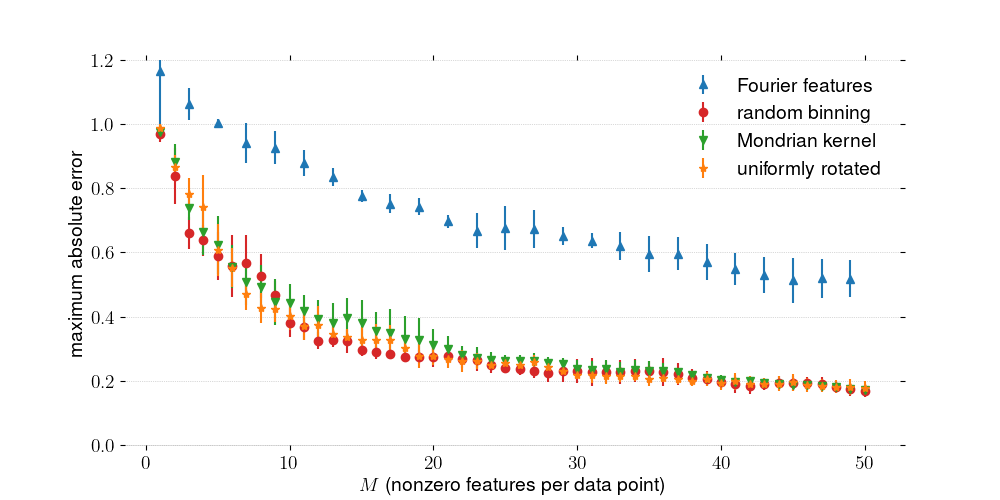}
         \caption{Convergence of Random Feature Kernel to Limiting Kernel for Uniform Points on $[0,1]^2$.\label{exp-1}}
     \end{subfigure}
     \begin{subfigure}[b]{0.49\textwidth}
         \centering
         \includegraphics[width=\textwidth]{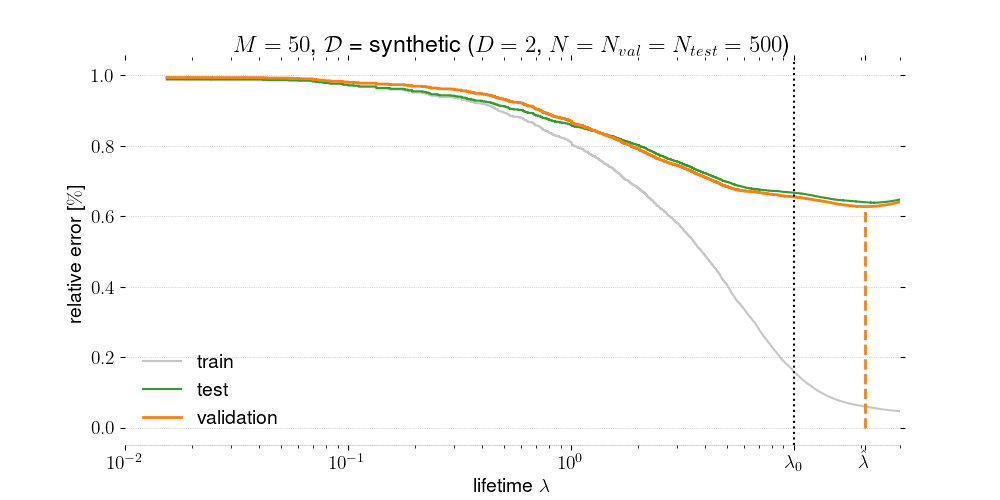}
         \caption{Recovery of Ground Truth Lifetime by Minimizing Validation Error for Synthetic Dataset.\label{exp-2}}
     \end{subfigure}
     \begin{subfigure}[b]{0.49\textwidth}
         \centering
         \includegraphics[width=\textwidth]{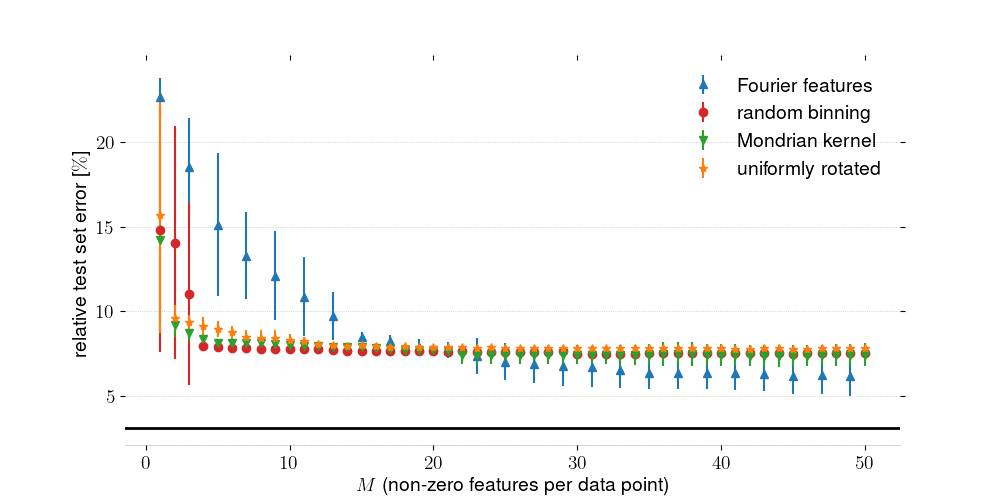}
         \caption{Validation Set Error versus Number of Nonzero Random Features for the \texttt{CPU} Dataset.\label{exp-3}}
     \end{subfigure}
     \begin{subfigure}[b]{0.49\textwidth}
         \centering
         \includegraphics[width=\textwidth]{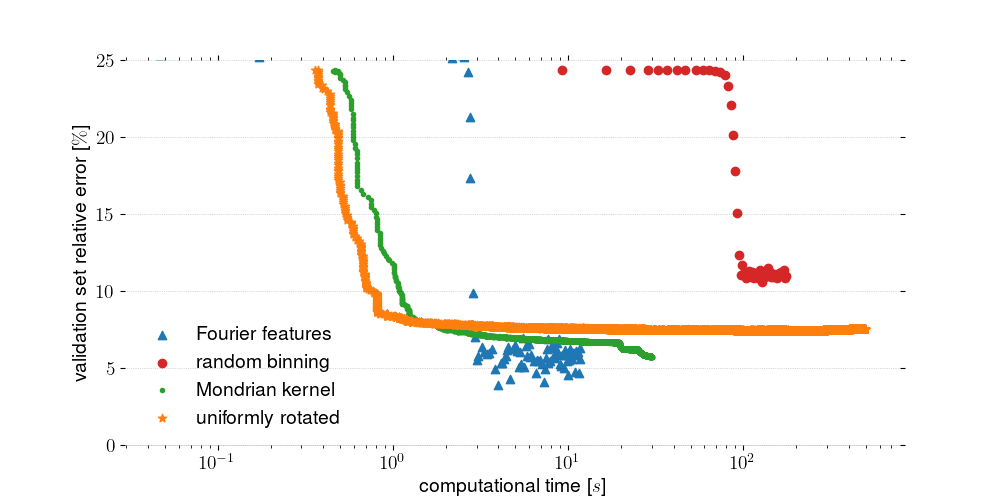}
         \caption{Validation Set Error versus Computational Time for the \texttt{CPU} Dataset.\label{exp-4}}
     \end{subfigure}
     \centering
     \begin{subfigure}[b]{0.49\textwidth}
         \includegraphics[width=\textwidth]{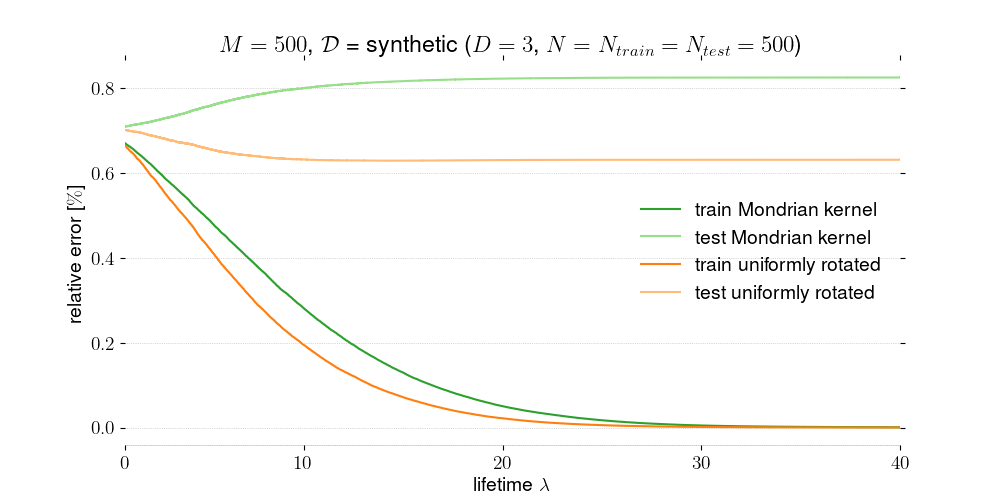}
         \caption{Relative Error versus Lifetime for the Mondrian Line Dataset.\label{exp-5}}
     \end{subfigure}
    \caption{Five Experiments Evaluating the Uniformly Rotated Mondrian Kernel in Regression Problems.}
\end{figure*}

To evaluate the empirical performance of the uniformly rotated Mondrian kernel, we ran five experiments, the first four of which are similar to those in \cite{Balog2016}, assessing the uniformly rotated Mondrian kernel against Fourier random features, random binning features, and the Mondrian kernel. 

\subsection{Experimental Procedures\label{procedures}}

In Experiment \subref{exp-1}, we demonstrate the convergence of each random feature map to its limiting kernel, namely, the Laplace kernel for Fourier random features, random binning features, and the Mondrian process, and the limiting kernel identified in Theorem \ref{thm:limiting_kernel} for the uniformly rotated Mondrian process. To do so, we uniformly generated $n = 100$ points in $[0,1]^2$ and plot the maximal absolute error between each random feature map and its limiting kernel over every pair of points for up to $M_{\max} = 50$ nonzero random features. For each random feature model, the lifetime parameter was set to $\lambda = 10$, and the experiment was repeated $\varepsilon = 5$ times to generate error bars.

In Experiment \subref{exp-2}, we demonstrate, as in \cite{Balog2016}, how the uniformly rotated Mondrian kernel can recover the ground truth lifetime of a synthetic dataset under an efficient kernel width selection procedure. To do so, we generate $N_{\text{train}} = N_{\text{test}} = N_{\text{validation}} = 500$ synthetic data points in $\mathbb{R}^2$ from the limiting kernel of the uniformly rotated Mondrian kernel with a lifetime parameter of $\lambda_0 = 10$. We then plot the train, test, and validation set errors from the uniformly rotated Mondrian kernel with $M = 50$ nonzero random features up to a maximum lifetime of $\lambda_{\text{max}} = 30$. We recover at the minimum validation set error a lifetime of $\hat{\lambda} \approx 21$, which is comparable to the recovered lifetime of $\hat{\lambda} \approx 19$ from an similar experiment on the Mondrian kernel from \cite{Balog2016} with the same ground truth lifetime.

In Experiment \subref{exp-3}, we compare the validation set errors of the random feature methods on the \texttt{CPU} dataset with respect to the number of nonzero random features, where, as in \cite{Balog2016}, the primal optimization problems were solved by stochastic gradient descent with a Ridge regularization constant of $\delta^2 = 10^{-4}$. To do so, for each random feature method, the validation set error was computed for up to $M_{\text{max}} = 50$ nonzero random features, and the experiment was repeated $\varepsilon = 5$ times to generate error bars. For Fourier random features, random binning features, and the Mondrian process, the lifetime parameter was set to $\lambda = 10^{-6}$ as in \cite{Balog2016}, and for the uniformly rotated Mondrian process, the lifetime parameter was set to $\lambda = 2.5 \times 10^{-7}$ based on the results of a parameter sweep. 

In Experiment \subref{exp-4}, we compare the validation set errors of the random feature methods on the \texttt{CPU} dataset with respect to the total computational time, where we again solve the primal optimization problems as in \cite{Balog2016}. To do so, for each random feature method, we used $M = 350$ nonzero random features. Furthermore, to set the lifetime parameter, a binary search algorithm as described in \cite{Balog2016} was implemented for Fourier random features and random binning features, while an efficient parameter sweep was used for the Mondrian kernel and uniformly rotated Mondrian kernel.

In Experiment \subref{exp-5}, we compare the relative training and testing set error of the Mondrian kernel and uniformly rotated Mondrian kernel on a synthetic dataset we call the \textit{Mondrian line}, which  is adapted from the Mondrian cube in \cite{TehRTFs2019}. To generate the Mondrian line, we first sample $N_{\text{train}} = N_{\text{test}} = 500$ datapoints uniformly on $[0, 1] \times [-\varepsilon, \varepsilon] \times [-\varepsilon, \varepsilon] \subset \mathbb{R}^3$ for a small value of $\varepsilon = 0.01$. Then, we label a datapoint $x_n$ as $1$ if either $x_n \in [0, 1] \times [0, \varepsilon] \times [0, \varepsilon]$ or $x_n \in [0, 1] \times [-\varepsilon, 0] \times [-\varepsilon, 0]$ and $0$ otherwise. Finally, we rotate each datapoint so that the dataset is biased toward the axis $\text{Span}(1, 1, 1)$. After constructing the dataset, we run Ridge regression with both the Mondrian kernel and uniformly rotated Mondrian kernel on the Mondrian line with $M = 500$ nonzero random features and plot the relative error with respect to the current lifetime.

\subsection{Discussion of Results}

In Experiment \subref{exp-1}, we confirm that the uniformly rotated Mondrian kernel indeed converges to an isotropic kernel, and, moreover, that it does so at a rate comparable to both random binning features and the Mondrian kernel, as well as faster than random Fourier features. In Experiment \subref{exp-2}, we confirm that the uniformly rotated Mondrian kernel maintains an essential feature of the Mondrian kernel, namely, the fast recovery of a ground truth lifetime. This allows the uniformly rotated Mondrian kernel to benefit from efficient hyperparamter selection as is described in \cite{Balog2016}.

In Experiment \subref{exp-3} and Experiment \subref{exp-4}, we see that the uniformly rotated Mondrian kernel performs comparably to the Mondrian kernel on the \texttt{CPU} data. This is particularly promising given the unique composition of the \texttt{CPU} dataset as an \textit{adversarial dataset} for the uniformly rotated Mondrian kernel. The \texttt{CPU} dataset has elements of the form $(x_1, \dots, x_d) \in \mathbb{R}^d$ for $d = 21$, with bounds on each dimension satisfying $0 \leq x_n < 10^4$ for all $n \not \in \{3, 8, 9, 20, 21\}$, $0 \leq x_n < 10^5$ for $n \in \{3, 20\}$, and $0 \leq x_n < 10^7$ for $n \in \{8, 9, 21\}$. From this observation, we see that the \texttt{CPU} dataset is stretched out across a comparably small number of standard coordinate axes. This suggests that the \texttt{CPU} dataset is particularly well-suited for the random feature methods that approximate the Laplace kernel, where the dependence on $||x - x'||_1$ differentiates well between points stretched along the $n \in \{8,9,21\}$ axes. Furthermore, this observation is a possible explanation as to why the uniformly rotated Mondrian kernel in Experiment \subref{exp-4} performs slightly worse in the long run on the \texttt{CPU} dataset compared to the Mondrian kernel, as the former's isotropic limiting kernel is less effective than the Laplace kernel on this particular dataset.

Furthermore, beyond this difference in limiting kernels, we see that the \texttt{CPU} dataset is particularly adversarial for the uniformly rotated Mondrian kernel with respect to \textit{computational intensity}, as generating a Mondrian tessellation on a particular uniform rotation of the dataset requires, on average, an initial bounding box of higher volume than that of the original dataset. In turn, this requirement demands a higher computational intensity for the uniformly rotated Mondrian kernel on the \texttt{CPU} dataset when compared to the Mondrian kernel. With these two factors in mind, it is promising that the uniformly rotated Mondrian kernel has comparable performance to the Mondrian kernel on the \texttt{CPU} dataset, a dataset that is well-suited for random feature models with a non-isotropic limiting kernel and, in particular, a dataset that is especially well-suited to the Mondrian kernel.

In Experiment \subref{exp-5}, we demonstrate, on a non-adversarial dataset, that the uniformly rotated Mondrian kernel can outperform the Mondrian kernel on the Mondrian line in both the training and testing relative error at all lifetimes $\lambda > 0$. While the \texttt{CPU} dataset is stretched out across a comparably small number of standard coordinate axes in $\mathbb{R}^{21}$, the Mondrian line is uniformly distributed across all standard coordinate axes in $\mathbb{R}^3$, making it particularly well-suited for random feature methods that approximate an isotropic kernel. For further experimental evidence that debiasing a model from a particular set of axes leads to improved empirical performance, see \cite{blaser2016random, TehRTFs2019, vempala:LIPIcs:2012:3847}.

In summary, beyond affirming the efficient convergence of the uniformly rotated Mondrian kernel to its limiting kernel and the recovery of a ground truth lifetime, our experiments demonstrate both that the uniformly rotated Mondrian kernel performs comparably to the Mondrian kernel on an adversarial dataset, and that the uniformly rotated Mondrian kernel outperforms the Mondrian kernel on a non-adversarial dataset.

\section{CONCLUSION}

The uniformly rotated Mondrian kernel as constructed in this work takes advantage of the computationally efficient Mondrian process to compute an isotropic kernel. The uniformly rotated Mondrian kernel converges uniformly to its limiting kernel at an exponential rate comparable to the Mondrian kernel. From our experiments, we observe on an adversarial dataset that the additional computational time spent on rotating the data to construct the uniformly rotated Mondrian kernel does not diminish the high efficiency of the Mondrian process, in addition to observing that the uniformly rotated Mondrian kernel can outperform the Mondrian kernel on a non-adversarial dataset. 

One extension of this work is to study an isotropic variant of \textit{Mondrian forests} as in \cite{lakshminarayanan2014mondrian}, a variation of the Mondrian kernel that is also based on generating $M$ independent Mondrian tessellations. As in our work above, we anticipate that a \textit{uniformly rotated Mondrian forest} would take advantage of the efficiency of the Mondrian process while approximating an isotropic kernel. For experiments studying Mondrian forests, see \cite{Balog2016}.

Finally, we expect that the lifting technique used in our proof of Equation \eqref{e:RZ_bnd} can be used to obtain more general results on the circumradius of the typical cell of Poisson and doubly stochastic Poisson hyperplane processes with discrete directional distributions, aiding the analysis of other transformed Mondrian estimators. Of particular interest is the case where the discrete directional distributions are informed by the training dataset, as such data-driven transformations could further improve the performance of the Mondrian kernel, e.g., the recent work of \cite{TrIM} with Mondrian forests.

\subsubsection*{Acknowledgments}

CO is grateful for support from the California Institute of Technology through the Summer Undergraduate Research Fellowship. EO is grateful for support from NSF Grant DMS-2402234.

\bibliography{refs}

\clearpage

\section*{Checklist}
 
\begin{enumerate}
    \item For all models and algorithms presented, check if you include:
    \begin{enumerate}
        \item A clear description of the mathematical setting, assumptions, algorithm, and/or model. [Yes] 
        \item[$\star$] \textit{See Section \ref{procedures} for complete details on our experiments.}
        
        \item An analysis of the properties and complexity (time, space, sample size) of any algorithm. [No] 
        \item[$\star$] \textit{Complexity analysis has not been done in either \cite{Balog2016} or \cite{RahimiRecht} for random feature models and would require extensive further work.}
        
        \item (Optional) Anonymized source code, with specification of all dependencies, including external libraries. [Yes] 
        \item[$\star$] \textit{See Section \ref{code} and} \texttt{README.md} \textit{for source code and dependencies.}
    \end{enumerate}

    \item For any theoretical claim, check if you include:
    \begin{enumerate}
        \item Statements of the full set of assumptions of all theoretical results. [Yes] 
        \item[$\star$] \textit{See the statement of Theorem \ref{thm:limiting_kernel}, Theorem \ref{thm:rate}, Proposition \ref{thm:lifting}, and Lemma \ref{thm:cells}, as well as the preceding paragraphs that explain all technical terminology.}
   
        \item Complete proofs of all theoretical results. [Yes] 
        \item[$\star$] \textit{See Section \ref{proofs}, Section \ref{proof_eq3}, Section \ref{proof_eq5}, and Section \ref{proof_thm2} for complete proofs of all theoretical results.}
   
        \item Clear explanations of any assumptions. [Yes] 
        \item[$\star$] \textit{The statement of all results include their full necessary assumptions.}
    \end{enumerate}

    \item For all figures and tables that present empirical results, check if you include:
    \begin{enumerate}
        \item The code, data, and instructions needed to reproduce the main experimental results (either in the supplemental material or as a URL). [Yes]  
        \item[$\star$] \textit{See Section \ref{code} and} \texttt{README.md}\textit{.}
   
        \item All the training details (e.g., data splits, hyperparameters, how they were chosen). [Yes] 
        \item[$\star$] \textit{See Section \ref{procedures}.}
    
        \item A clear definition of the specific measure or statistics and error bars (e.g., with respect to the random seed after running experiments multiple times). [Yes] 
        \item[$\star$] \textit{See Section \ref{code},} \texttt{README.md}\textit{, and Section \ref{procedures}.}

        \item A description of the computing infrastructure used. (e.g., type of GPUs, internal cluster, or cloud provider). [Not Applicable]
    \end{enumerate}

    \item If you are using existing assets (e.g., code, data, models) or curating/releasing new assets, check if you include:
    \begin{enumerate}
        \item Citations of the creator if your work uses existing assets. [Yes] 
        \item[$\star$] \textit{Our figures are all independently created, and the framework of our code from \cite{Balog2016} is cited with its original license in} \texttt{NOTICE.md} \textit{and with further detail in} \texttt{README.md}\textit{.}
   
        \item The license information of the assets, if applicable. [Yes]
        \item[$\star$] \textit{The license information of the framework of our code from \cite{Balog2016} is included in } \texttt{NOTICE.md} \textit{and further detailed in} \texttt{README.md}\textit{.}
   
        \item New assets either in the supplemental material or as a URL, if applicable. [Not Applicable]
   
        \item Information about consent from data providers/curators. [Not Applicable] 
   
        \item Discussion of sensible content if applicable, e.g., personally identifiable information or offensive content. [Not Applicable]
    \end{enumerate}
    
    \item If you used crowdsourcing or conducted research with human subjects, check if you include:
    \begin{enumerate}
        \item The full text of instructions given to participants and screenshots. [Not Applicable]
        
        \item Descriptions of potential participant risks, with links to Institutional Review Board (IRB) approvals if applicable. [Not Applicable]
        \item The estimated hourly wage paid to participants and the total amount spent on participant compensation. [Not Applicable]
    \end{enumerate}
\end{enumerate}
 
\onecolumn
\thispagestyle{empty}

\aistatstitle{The Uniformly Rotated Mondrian Kernel \\
Supplementary Materials}


\appendix
\section{APPENDIX}

In this Appendix, we will provide detailed proofs of our theoretical results from above.

\subsection{Proof of Equation \eqref{e:vZ_bnds}\label{proof_eq3}}

\textit{Proof of Equation \eqref{e:vZ_bnds}.} 
Let $X$ be the superposition of $M$ independent rotated Poisson Manhattan tessellations $X_1, \dots, X_M$ all with intensity $\lambda > 0$. For $n=1, \ldots, M$, note that the Poisson Manhattan tessellation $X_n$ has directional distribution
\begin{equation*}
    \varphi_n = \dfrac{1}{d}\sum_{i=1}^d \delta_{u_{n,i}}
\end{equation*}
for some orthonormal set of vectors $\{u_{n,1}, \dots, u_{n,d}\}$, and thus the associated zonoid of $X_n$ has support function

\begin{equation*}
    h_{\Pi_{X_n}}(u) = \dfrac{\lambda}{2} \int_{S^{d-1}} \left|\left\langle u, v\right\rangle \right|\mathrm{d}\varphi_n(v) = \left(\dfrac{\lambda}{2d}\right) \sum_{i=1}^d \left|\left\langle u, u_{n,i}\right\rangle \right|.
\end{equation*}
The above quantity achieves the same maximum and minimum values over the unit sphere $||u||_{2} = 1$ as 
\begin{equation*}
    \left(\dfrac{\lambda}{2d}\right) \sum_{i=1}^d \left|\left\langle u, e_i\right\rangle \right| = \left(\dfrac{\lambda}{2d}\right)||u||_{1}.
\end{equation*}
Thus, by the inequality $||u||_2 \leq ||u||_1 \leq \sqrt{d}||u||_2$ that is tight over the unit sphere $||u||_{2} = 1$, we have that
\begin{equation*}
    \min_{u \in S^{d-1}} h_{\Pi_{X_n}}(u) = \frac{\lambda}{2d}, \quad \max_{u \in S^{d-1}} h_{\Pi_{X_n}}(u) = \dfrac{\lambda}{2 \sqrt{d}}.
\end{equation*}
From Page 158 of \cite{Schneider2008}, we have that the associated zonoid of the superposition $X$ of the independent stationary Poisson hyperplane processes $X_1, \ldots, X_M$ is additive in the sense that
\begin{align*}
        \Pi_{X} =  \sum_{n=1}^M \Pi_{X_n},
\end{align*}
where the associated sum is the Minkowski sum of the associated zonoids, i.e.,  we have that $\Pi_X = \left\{v_1 + \dots + v_M : v_n \in \Pi_{X_n}, 1 \leq n \leq M \right\}$. Furthermore, as the support function of the Minkowski sum of convex bodies is additive, it follows that
\begin{align*}
    R(\Pi_{X}) &\leq \max_{u \in S^{d-1}} h_{\Pi_{X}}(u) = \max_{u \in S^{d-1}} \sum_{n=1}^M h_{\Pi_{X_n}}(u) \leq \sum_{n=1}^M \max_{u \in S^{d-1}} h_{\Pi_{X_n}}(u) = \dfrac{\lambda M}{2\sqrt{d}}, \\
    r(\Pi_{X}) &\geq \min_{u \in S^{d-1}} h_{\Pi_{X}}(u) = \min_{u \in S^{d-1}} \sum_{n=1}^M h_{\Pi_{X_n}}(u) \geq \sum_{n=1}^M \min_{u \in S^{d-1}} h_{\Pi_{X_n}}(u) = \dfrac{\lambda M}{2d}.
\end{align*}
From Theorem 10.3.3 and (10.4) in \cite{Schneider2008}, it holds that $\mathbb{E}[V(Z)] = V(\Pi_X)^{-1}$. 
\vfill
\newpage
Thus, we have that
\begin{align*}
    \mathbb{E}\left[V(Z)\right] &\leq \left(\kappa_d \cdot r(\Pi_{X})^d\right)^{-1} \leq \dfrac{1}{\kappa_d}\left(\dfrac{2d}{\lambda M}\right)^d, \\
    \mathbb{E}\left[V(Z)\right] &\geq \left(\kappa_d \cdot R(\Pi_{X})^d\right)^{-1} \geq \dfrac{1}{\kappa_d}\left(\dfrac{2\sqrt{d}}{\lambda M}\right)^d.
\end{align*}
This completes our computation of bounds for the expected volume of the typical cell of the superposition of independent rotated Mondrian tessellations. \hfill $\square$

\subsection{Proof of Equation \eqref{e:RZ_bnd}\label{proof_eq5}}

To prove Equation \eqref{e:RZ_bnd}, we will first provide an explicit lift for the superposition of independent rotated Mondrian tessellations. While similar to Theorem 3.1 in \cite{OReilly2022} in providing a lift for a STIT tessellation with finitely many cut directions, our result below gives both an explicit characterization of the subspace $U$ and the Poisson Manhattan tessellation $\widetilde{X}$ in the lifted space for our specific setting.

\begin{proposition}\label{thm:lifting} 
Let $X$ be the superposition of $M$ independent rotated Poisson Manhattan tessellations $X_1, \dots, X_M$ all with intensity $\lambda > 0$ in $\mathbb{R}^d$, where each $X_n$ has directional distribution
\begin{equation*}
    \varphi_n = \dfrac{1}{d} \sum_{i=1}^d \delta_{u_{n,i}}
\end{equation*}
for an orthonormal set of vectors $\{u_{n,1}, \dots, u_{n,d}\}$. It holds that $X$ is the same in distribution as $\widetilde{X} \cap U$, where $\widetilde{X}$ is an axis-aligned Poisson Manhattan tessellation in $\mathbb{R}^{Md}$ with intensity $\widetilde{\lambda} = M^{3/2}\lambda > 0$ and $U = \text{Im}(T)$ is a $d$-dimensional linear subspace of $\mathbb{R}^{Md}$, where the orthogonal transformation $T : \mathbb{R}^{d} \to \mathbb{R}^{Md}$ is represented by the matrix
\begin{equation}\label{e:T}
    \mathcal{M}(T) = \dfrac{1}{\sqrt{M}} \begin{bmatrix}
        | & & | & & | & & | \\
        u_{1,1} & \dots & u_{1,d} & \dots & u_{M, 1} & \dots & u_{M, d} \\
        | & & | & & | & & | 
    \end{bmatrix}^T. 
\end{equation}
\end{proposition}

\noindent \textit{Proof of Proposition \ref{thm:lifting}:} First, we recall from Page 158 of \cite{Schneider2008} that the directional distribution of $X$ is given by
\begin{equation*}
    \varphi = \dfrac{1}{M}\sum_{n=1}^M \varphi_n = \dfrac{1}{Md} \sum_{n=1}^M \sum_{i=1}^d \delta_{u_{n,i}}.
\end{equation*}
We then have from $(4.60)$ in \cite{Schneider2008} that the intensity of the point process $X \cap \text{Span}(u)$ for any $u \in S^{d-1}$ is given by twice the support function $h_{\Pi_X}(u)$ of the associated zonoid $\Pi_X$ of $X$, i.e., we have that
\begin{equation*}
    \lambda_{X \cap \text{Span}(v)} = M\lambda \int_{S^{d-1}} |\langle u, v \rangle| \mathrm{d}\varphi(v) = \dfrac{\lambda }{d} \sum_{n=1}^M \sum_{i=1}^d |\langle u, u_{n,i} \rangle|
\end{equation*}
as in our computation in proving Equation \eqref{e:vZ_bnds}. Similarly, for any $\widetilde{u} \in S^{Md - 1}$, recalling that
\begin{equation*}
    \widetilde{\varphi} = \dfrac{1}{Md} \sum_{k=1}^{Md} \delta_{e_k}
\end{equation*}
is the directional distribution of $\widetilde{X}$, we have that
\begin{equation*}
    \lambda_{\widetilde{X} \cap \text{Span}(\widetilde{u})} = M^{3/2}\lambda \int_{S^{Md-1}} |\langle \widetilde{u}, v\rangle|\mathrm{d} \widetilde{\varphi}(v) = \dfrac{\lambda\sqrt{M}}{d} \sum_{k=1}^{Md} |\langle \widetilde{u}, e_k\rangle| = \dfrac{\lambda\sqrt{M}}{d} ||\widetilde{u}||_1.
\end{equation*}
We will then consider the linear transformation $T : \mathbb{R}^d \to \mathbb{R}^{Md}$ as in \eqref{e:T}. We note for any $u \in \mathbb{R}^d$ that
\begin{equation*}
    Tu = \dfrac{1}{\sqrt{M}}\left(\left\langle u, u_{1,1} \right\rangle, \,\dots\,, \left\langle u, u_{1,d} \right\rangle, \,\dots\,, \left\langle u, u_{M,1} \right\rangle, \,\dots\,, \left\langle u, u_{M,d} \right\rangle\right),
\end{equation*}
and so we have that
\begin{align*}
    \lambda_{X \cap \text{Span}(u)} = \dfrac{\lambda}{d} \sum_{i=1}^M \sum_{n=1}^d \left|\left\langle u, u_{n, i}\right\rangle\right| = \dfrac{\lambda \sqrt{M}}{d} \left|\left|Tu\right|\right|_{1}= \lambda_{\widetilde{X} \cap \text{Span}(Tu)}.
\end{align*}
By uniqueness on Page 131 of \cite{Schneider2008}, it follows that $X$ is distributed the same as $\widetilde{X} \cap U$, where $U = \text{Im}(T)$. Finally, we note since each $\{u_{n, 1}, \dots, u_{n, d}\}$ is an orthonormal set of vectors, the matrix
\begin{align*}
    \begin{bmatrix}
        | &\dots  & |\\
        u_{n, 1} & \dots & u_{n, d} \\
        | & \dots & |
    \end{bmatrix}^T &= \begin{bmatrix}
        \text{—} & u_{n, 1} & \text{—} \\
        \vdots & \vdots & \vdots \\
        \text{—} & u_{n, d} & \text{—}
    \end{bmatrix}
\end{align*}
is orthogonal and 
\begin{align*}
    \left\langle (u_{n,1}^{(j_1)}, \dots, u_{n,d}^{(j_1)}), (u_{n,1}^{(j_2)}, \dots, u_{n,d}^{(j_2)}) \right\rangle  = \begin{cases}
        1, &j_1 = j_2, \\
        0,&\text{otherwise},
    \end{cases}
\end{align*}
where $u_{n,i} = (u_{n,i}^{(1)}, \dots, u_{n,i}^{(d)})$ are the coordinates of the vector $u_{n,i}$. We thus have for all $1 \leq j_1, j_2 \leq d$ that
\begin{align*}
    &\left\langle T e_{j_1}, T e_{j_2} \right\rangle \\
    &= \dfrac{1}{M}\left(u_{1,1}^{(j_1)} u_{1,1}^{(j_2)} + \dots + u_{1,d}^{(j_1)} u_{1,d}^{(j_2)} + \dots + u_{M,1}^{(j_1)}u_{M,1}^{(j_2)} + \dots + u_{M,d}^{(j_1)} u_{M,d}^{(j_2)}\right) \\[1em]
    &= \dfrac{1}{M}\left(\left\langle(u_{1,1}^{(j_1)}, \dots, u_{1,d}^{(j_1)}), (u_{1,1}^{(j_2)}, \dots, u_{1,d}^{(j_2)})\right\rangle + \dots + \left\langle (u_{M,1}^{(j_1)}, \dots, u_{M,d}^{(j_1)}), (u_{M,1}^{(j_2)}, \dots, u_{M,d}^{(j_2)})\right\rangle \right) \\[1em]
    &= \begin{cases}
        1, &j_1 = j_2 \\
        0, &\text{otherwise}.
    \end{cases}
\end{align*}
We thus have that $T : \mathbb{R}^d \to \mathbb{R}^{Md}$ is an orthogonal transformation, and so $U$ is a $d$-dimensional linear subspace of $\mathbb{R}^{Md}$ such that $T$ preserves the structure of $X$ as a random tessellation as is desired. \hfill $\square$ \\

\noindent \textit{Proof of Equation \eqref{e:RZ_bnd}}. We will let $\widetilde{X}$ and $U = \text{Im}(T)$ be as defined in Proposition \ref{thm:lifting} for $X$ a superposition of $M$ independent rotated Poisson Manhattan tessellations $X_1, \dots, X_M$ so that $X$ and $\widetilde{X} \cap U$ are the same in distribution. We have in particular for the typical cell $Z$ of $X$ that
\begin{align*}
    \mathbb{P}[R(Z) \geq a] = \mathbb{P}\left[R\left(U \cap \widetilde{Z}\right) \geq a\right],
\end{align*}
where the typical cell $\widetilde{Z}$ of $\widetilde{X}$ is distributed as 
\begin{equation*}
    \prod_{k=1}^{Md} [-t_k, t_k].
\end{equation*}
We recall that the particular orthogonal lift $T : \mathbb{R}^d \to \mathbb{R}^{Md}$ of $X$ is represented by the matrix
\begin{equation*}
    \mathcal{M}(T) = \begin{bmatrix}
        u_1, \dots, u_d
    \end{bmatrix} = \dfrac{1}{\sqrt{M}}\begin{bmatrix}
        \mid & & \mid & & \mid & & \mid \\
        u_{1,1} & \dots & u_{1,d} & \dots & u_{M,1} & \dots & u_{M,d} \\
        \mid & & \mid & & \mid & & \mid
    \end{bmatrix}^T
\end{equation*}
for orthonormal sets of vectors $\{u_{1,i}, \dots, u_{d,i}\}$ with each $u_{n,i} \in \mathbb{R}^d$ for all $1 \leq i \leq M$. Furthermore, we have for all fixed $t_1, \dots, t_{Md} \in (0, \infty)$ that
\begin{align*}
    R\left(U \cap \prod_{k=1}^{Md}[-t_k, t_k]\right) &= \max\left\{\left|\left|x\right|\right|_{2} : x \in U \cap \prod_{k=1}^{Md}[-t_k, t_k]\right\} \\[1em]
    &= \max\left\{\left|\left|c\right|\right|_{2} : \sum_{i=1}^d c_iu_i \in \prod_{k=1}^{Md}[-t_k, t_k]\right\},
\end{align*}
where we recall since $T$ is orthogonal that $u_1, \dots, u_d \in \mathbb{R}^{Md}$ is orthogonal. We will then define
\begin{equation*}
    \begin{bmatrix}
        \mid & & \mid \\
        u_{n,1} & \dots & u_{n,d} \\
        \mid & & \mid 
    \end{bmatrix}^T = \begin{bmatrix}
        \mid & & \mid \\
        v_{n,1} & \dots & v_{n,d} \\
        \mid & & \mid 
    \end{bmatrix}
\end{equation*}
so that
\begin{align*}
    \mathcal{M}(T) = \dfrac{1}{\sqrt{M}}\begin{bmatrix}
        \mid & & \mid & & \mid & & \mid \\
        u_{1,1} & \dots & u_{1,d} & \dots & u_{M,1} & \dots & u_{M,d} \\
        \mid & & \mid & & \mid & & \mid
    \end{bmatrix}^T &= \dfrac{1}{\sqrt{M}} \begin{bmatrix}
        \mid & & \mid \\
        v_{1,1} & \dots & v_{1,d} \\
        \mid & & \mid \\
        \vdots & \vdots & \vdots \\
        \mid & & \mid \\
        v_{M,1} & \dots & v_{M,d} \\
        \mid & & \mid
    \end{bmatrix} \\[1em]
    &= \begin{bmatrix}
        u_1 & \dots & u_d
    \end{bmatrix},
\end{align*}
where each $v_{n,1}, \dots, v_{n,d} \in \mathbb{R}^d$ is orthonormal. It follows for any
\begin{equation*}
    \sum_{i=1}^d c_iu_i \in \prod_{k=1}^{Md} [-t_k, t_k],
\end{equation*}
that 
\begin{align*}
    \sum_{i=1}^dc_iu_i^{(k)} &\in [-t_k, t_k]\text{ for all }1 \leq k \leq Md \\
    &\Rightarrow \sum_{i=1}^dc_i\left(\dfrac{v_{n,i}}{\sqrt{M}}\right) \in \prod_{k = nd + 1}^{(n + 1)d}[-t_k, t_k]\text{ for all }1 \leq n \leq M \\[1em]
    &\Rightarrow \left|\left|\sum_{i=1}^dc_i\left(\dfrac{v_{n,i}}{\sqrt{M}}\right)\right|\right|_{2} \overset{(\dagger)}{=} \dfrac{1}{\sqrt{M}} ||c||_{2}\leq \sqrt{t_{nd + 1}^2 + \dots + t_{(n+1)d}^2} \text{ for all }1 \leq n \leq M, 
\end{align*}
where $(\dagger)$ holds since each set $\{v_{n,1}, \dots, v_{n,d}\}$ is orthogonal. We have thus computed an upper bound of the form
\begin{align*}
    R\left(U \cap \prod_{k=1}^{Md}[-t_k, t_k]\right) &= \max\left\{\left|\left|c\right|\right|_{2} : \sum_{n=1}^d c_nu_n \in \prod_{n=1}^{Md}[-t_k, t_k]\right\} \\[1em]
    &\leq \text{min}\left(\sqrt{t_1^2 + \dots + t_d^2}, \dots, \sqrt{t_{(M-1)d + 1}^2 + \dots + t_{Md}^2}\right)\sqrt{M} \\[1em]
    &\leq \text{min}\left(t_1 + \dots + t_d, \dots, t_{(M-1)d + 1} + \dots + t_{Md}\right)\sqrt{M}
\end{align*}
for any particular set of rotated Poisson Manhattan tessellations $X_1, \dots, X_M$. Finally, since $\widetilde{X}$ is a Poisson Manhattan tessellation in $\mathbb{R}^{Md}$ of intensity $\lambda M^{3/2} > 0$, we know that $t_k \sim \text{Exp}(2 \lambda \sqrt{M}/d)$. It follows that each $t_k\sqrt{M} \sim \text{Exp}(2\lambda /d)$ so that 
\begin{align*}
    (t_{id + 1} + \dots + t_{(i + 1)d}) &\sqrt{M} \sim \text{Erlang}\left(d, \dfrac{2 \lambda}{d}\right), \\
    &\text{where } \mathbb{P}\left[\text{Erlang}\left(d, \dfrac{2 \lambda}{d}\right) \geq a\right] = \exp\left(-\dfrac{2 \lambda a}{d}\right)\sum_{n=0}^{d-1} \dfrac{1}{n!}\left(\dfrac{2 \lambda a}{d}\right)^n.
\end{align*}
Finally, since for all independent and identically distributed random variables $Y_1, \dots, Y_n$ it holds that
\begin{equation*}
    \mathbb{P}[\text{min}(Y_1, \dots, Y_n) \geq a] = \mathbb{P}[Y_1 \geq a]^n,
\end{equation*}
we have that
\begin{equation*}
    \mathbb{P}\left[R(Z) \geq a\right] \leq \exp\left(-\dfrac{2 M \lambda  a}{d}\right)\left(\sum_{n=0}^{d-1} \dfrac{1}{n!}\left[\dfrac{2 \lambda a}{d}\right]^n\right)^M
\end{equation*}
as is desired, completing our computation of an upper bound for the circumradius of the typical cell of the superposition of independent rotated Mondrian tessellations. \hfill $\square$

\subsection{Proof of Theorem \ref{thm:rate} \label{proof_thm2}}

To prove Theorem \ref{thm:rate}, we will need to make use of the application of Campbell's theorem that was stated above.

\setcounter{theorem}{2}
\begin{theorem} \textup{\citep[Equation 4.3]{Schneider2008}} 
For any nonnegative measurable function $f$ on the space of nonempty compact sets in $\mathbb{R}^d$ and for any stationary random tessellation $X$, we have that
\begin{equation*}
    \mathbb{E}\left[\sum_{C \in \text{cells}(X)}f(C)\right] = \dfrac{1}{\mathbb{E}[V(Z)]} \cdot \mathbb{E}\left[\int_{\mathbb{R}^d} f(Z + y)\mathrm{d}y\right],
\end{equation*}
where $Z$ is the typical cell of the $X$.
\end{theorem}

\textit{Proof of Theorem \ref{thm:rate}.} We will consider the definitions and events as are detailed in the outline from Section \ref{proofs}. From $A_1 \cap A_2 \cap A_3$, we have for every $x \in \mathcal{X}$ that there is a point $u \in \mathcal{U}$ such that $x,u$ are in the same cell of $X$ and
\begin{equation*}
    \left|\left| x - u\right|\right|_{2} \leq \dfrac{\delta}{4 \lambda \sqrt{d}}.
\end{equation*}
Then, for any $x,x' \in \mathcal{X}$ we have that
\begin{align*}
    \left|k_M(x, x') - k_\infty(x, x')\right| \leq \left| k_M(x, x') - k_M(u, u')\right| + \left|k_M(u, u') - k_\infty(u, u')\right| + \left|k_\infty(u, u') - k_\infty(x, x')\right|,
\end{align*}
where $u,u' \in \mathcal{U}$ are as described above. Since $x,u$ and $x',u'$ are in the same cells of $X$,
\begin{equation*}
    k_M(x, x') = k_M(u, u') \Rightarrow \left| k_M(x, x') - k_M(u, u')\right| = 0
\end{equation*}
for all $x,x' \in \mathcal{X}$. We then note from $A_4$ that $\sup_{x,x' \in \mathcal{X}} \left| k_M(u, u') - k_\infty(u, u') \right| \leq \delta/2$. We finally note from Theorem \ref{thm:limiting_kernel} that
\begin{align*}
    \sup_{x,x' \in \mathcal{X}} \left|k_\infty(u, u') - k_\infty(x, x')\right| &= \sup_{x,x' \in \mathcal{X}} \dfrac{1}{\omega_d}\left|\int_{S^{d-1}}\left(e^{-\lambda \left| \left|u - u'\right|\right|_{2} \left|\left|v\right|\right|_{1}} - e^{-\lambda \left| \left|x - x'\right|\right|_{2} \left|\left|v\right|\right|_{1}}\right)\mathrm{d}v \right| \\[1em]
    &\leq \sup_{x,x' \in \mathcal{X}}\dfrac{1}{\omega_d}\int_{S^{d-1}}\left|e^{-\lambda\left|\left|u - u'\right|\right|_{2} \left|\left|v\right|\right|_{1}} - e^{-\lambda\left|\left|x - x'\right|\right|_{2}\left|\left|v\right|\right|_{1}}\right|\mathrm{d}v  \\[1em]
    &\overset{(\mathrm{a})}{\leq} \sup_{x,x' \in \mathcal{X}}\lambda\left|\left|\left|u - u'\right|\right|_{2} - \left|\left|x - x'\right|\right|_{2}\right| \dfrac{1}{\omega_d}\int_{S^{d-1}}\left|\left|v\right|\right|_{1}\mathrm{d}v \\[1em]
    &\overset{(\mathrm{b})}{\leq} \sup_{x,x' \in \mathcal{X}}\lambda \sqrt{d}\left|\left|\left|u - u'\right|\right|_{2} - \left|\left|x - x'\right|\right|_{2}\right| \dfrac{1}{\omega_d}\int_{S^{d-1}}\left|\left|v\right|\right|_{2}\mathrm{d}v\\[1em]
    &\overset{(\mathrm{c})}{=} \sup_{x,x' \in \mathcal{X}}\lambda \sqrt{d}\left|\left|\left|u - u'\right|\right|_{2} - \left|\left|x - x'\right|\right|_{2}\right|\\[1em]
    &\leq\sup_{x,x' \in \mathcal{X}} \lambda \sqrt{d} \left|\left|(u - u') - (x - x')\right|\right|_{2} \\[1em]
    &\leq \sup_{x,x' \in \mathcal{X}}\lambda \sqrt{d} \left(\left|\left|u - x\right|\right|_{2} + \left|\left|u' - x'\right|\right|_{2}\right) \leq \dfrac{\delta}{2},
\end{align*}
where $(\mathrm{a})$ follows from the inequality $\left|e^{-|x|} - e^{-|y|}\right| \leq \left|x - y\right|,$  $(\mathrm{b})$ follows from the inequality $\left|\left|v\right|\right|_{1} \leq \sqrt{d} \left|\left|v\right|\right|_{2}$, and $(\mathrm{c})$ holds from
\begin{equation*}
    \dfrac{1}{\omega_d}\int_{S^{d-1}} \left|\left|v\right|\right|_{2}\mathrm{d}v = \dfrac{1}{\omega_d}\int_{S^{d-1}} \mathrm{d}v = 1.
\end{equation*}
We thus have that the $\delta$-approximation on $\mathcal{X}$ is satisfied on the event $A_1 \cap A_2 \cap A_3 \cap A_4$, and so
\begin{align*}
    \mathbb{P}\left[\sup_{x,x' \in \mathcal{X}}\left|k_M(x, x') - k_\infty(x, x')\right| > \delta\right] &\leq \mathbb{P}(A_1^c \cup A_2^c \cup A_3^c \cup A_4^c) \\[1em]
    &\leq \mathbb{P}(A_1^c) + \mathbb{P}(A_2^c) + \mathbb{P}(A_3^c) + \mathbb{P}(A_4^c),
\end{align*}
from a union bound. We will now bound these four probabilities. First, we saw from the outline above that
\begin{equation*}
    \mathbb{P}(A_1^c) \leq \left(\dfrac{\kappa_d^2 d}{2 \lambda M}\right)\left(\dfrac{\lambda M}{2 \sqrt{d}}\right)^d e^{-2 \lambda M (R - r)}.
\end{equation*}
Second, we see that for $A_2^c$,
\begin{align*}
    \mathbb{P}(A_2^c) &= \mathbb{P}[\text{there exists a cell in }X\text{ with circumcenter in }B_R\text{ disjoint from }\mathcal{U}] \\[1em]
    &= \mathbb{P}\left[\bigcup_{\text{cells }C \in X}(c(C) \in B_R) \cap (C\text{ is disjoint from }\mathcal{U})\right] \\[1em]
    &\leq \mathbb{E}\left[\sum_{\text{cells }C \in X} \mathds{1}[c(C) \in B_R]\mathds{1}\left[C\text{ is disjoint from }\mathcal{U}\right]\right] \\[1em]
    &= \left(\dfrac{1}{\mathbb{E}[V(Z)]}\right) \mathbb{E}\left[\int_{\mathbb{R}^d}\mathds{1}[y \in B_R]\mathds{1}\left[Z+y\text{ is disjoint from }\mathcal{U}\right]\mathrm{d}y\right],
\end{align*}
where we have again used Campbell's Theorem in the last equality. Since $r(C) \geq \varepsilon/2 \sqrt{d}$ implies that $C$ contains a point of $\mathcal{U}$, we note that
\begin{align*}
    \mathbb{P}(A_2^c) \leq \left(\dfrac{1}{\mathbb{E}[V(Z)]}\right) \mathbb{E}\left[\int_{\mathbb{R}^d} \mathds{1}[y \in B_R] \mathds{1}[r(Z + y) < \varepsilon/2 \sqrt{d}]\mathrm{d}y\right] &= \left(\dfrac{\kappa_d R^d}{\mathbb{E}[V(Z)]}\right) \mathbb{P}[r(Z) < \varepsilon/2 \sqrt{d}] \\[1em]
    &\leq \kappa_d^2\left(\dfrac{\lambda R M}{2\sqrt{d}}\right)^d\mathbb{P}[r(Z) < \varepsilon/2 \sqrt{d}] \\[1em]
    &= \kappa_d^2\left(\dfrac{\lambda R M}{2\sqrt{d}}\right)^d\left(1 - e^{- M \lambda \varepsilon \sqrt{d}}\right) \\[1em]
    &\leq \kappa_d^2\left(\dfrac{\lambda R M}{2\sqrt{d}}\right)^d M \lambda \varepsilon \sqrt{d},
\end{align*}
where in the last inequality we have used the fact that $1 - e^{-x} \leq x$ for all $x\in \mathbb{R}$.

Third, we see that for $A_3^c$,
\begin{align*}
    \mathbb{P}(A_3^c) &= \mathbb{P}[\text{there exists a cell in }X\text{ with circumcenter in }B_R\text{ and diameter } \geq \delta/4\lambda \sqrt{d}] \nonumber\\[1em]
    &= \mathbb{P}\left[\bigcup_{\text{cells }C \in X} \left(c(C) \in B_R\right)\cap \left(R(C) \geq \dfrac{\delta}{8 \lambda \sqrt{d}}\right)\right] \\[1em]
    &\leq \sum_{\text{cells }C \in X} \mathbb{P}\left[c(C) \in B_R\right]\mathbb{P}\left[R(C) \geq \dfrac{\delta}{8 \lambda \sqrt{d}}\right] \\[1em]
    &= \mathbb{E}\left[\sum_{\text{cells }C \in X}\mathds{1}\left[c(C) \in B_R\right] \mathds{1}\left[R(C) \geq \dfrac{\delta}{8 \lambda \sqrt{d}}\right]\right] \\[1em]
    &= \left(\dfrac{1}{\mathbb{E}\left[V(Z)\right]}\right)\mathbb{E}\left[\int_{\mathbb{R}^d}\mathds{1}\left[y \in B_R\right] \mathds{1}\left[R(Z + y) \geq \dfrac{\delta}{8 \lambda \sqrt{d}}\right]\mathrm{d}y\right] \\[1em]
    &= \left(\dfrac{\kappa_d R^d}{\mathbb{E}\left[V(Z)\right]}\right) \mathbb{P}\left[R(Z) \geq \dfrac{\delta}{8 \lambda \sqrt{d}}\right],
\end{align*}
with $R(C + y) = R(C)$ for all cells $C$ since the circumradius is translation invariant. We thus have from \eqref{e:RZ_bnd} that
\begin{equation*}
    \mathbb{P}(A_3^c) \leq \kappa_d^2\left(\dfrac{\lambda RM}{2\sqrt{d}}\right)^d \left(\sum_{n=0}^{d-1} \dfrac{1}{n!}\left[ \dfrac{\delta}{4 d^{3/2}} \right]^n\right)^Me^{-M \delta/4d^{3/2}}.
\end{equation*}
Finally, we see that for $A_4^c$ by Hoeffding's inequality that
\begin{align*}
    \mathbb{P}(A_4^c) = \mathbb{P}[\text{the }\delta/2\text{-approximation fails on }\mathcal{U}] &= \mathbb{P}\left[\text{there exists }u,u' \in \mathcal{U}\text{ such that }\left|k_M(u,u') - k_\infty(u,u')\right| > \delta/2\right] \\[1em]
    &= \mathbb{P}\left[\bigcup_{u,u' \in \mathcal{U}}\left|k_M(u,u') - k_\infty(u,u')\right| > \delta/2\right] \\[1em]
    &\leq \sum_{u,u' \in \mathcal{U}} \mathbb{P}\left[\left|k_M(u,u') - k_\infty(u,u')\right| > \delta/2\right] \leq \sum_{u,u' \in \mathcal{U}} 2 \exp\left(-M\delta^2/2\right).
\end{align*}
For any $\varepsilon$-grid covering $\mathcal{U}$ of $B_{R+\varepsilon}$, we have that
\begin{equation*}
    \#(\mathcal{U}) \leq \prod_{n=1}^d \left\lceil\dfrac{2[R+\varepsilon]}{\varepsilon}\right\rceil \leq  \prod_{n=1}^d \left(\dfrac{4 [R+\varepsilon]}{\varepsilon}\right) = \left(\dfrac{4R}{\varepsilon} + 1\right)^d
\end{equation*}
since $\varepsilon < 2r$. We thus have that
\begin{align*}
    \mathbb{P}(A_4^c) \leq \left[\left(\dfrac{4R}{\varepsilon} + 1\right)^d\right]^2 \left(2e^{-M\delta^2/2}\right) &= \left(2e^{-M\delta^2/2}\right) \sum_{n=0}^{2d}\binom{2d}{n}\left(\dfrac{4R}{\varepsilon}\right)^n < \left(2e^{-M\delta^2/2}\right) \left(\sum_{n=0}^{2d}\binom{2d}{n}(4R)^n \right)\varepsilon^{-2d}
\end{align*}
for all small enough $0 < \varepsilon < 1$. This gives us that
\begin{align*}
    \mathbb{P}\left[\sup_{x, x' \in \mathcal{X}}\left|k_M(x, x') - k_\infty(x, x')\right| > \delta\right] < &\left(\dfrac{\kappa_d^2 d}{2 \lambda M}\right)\left(\dfrac{\lambda M}{2 \sqrt{d}}\right)^d e^{-2 \lambda M (R - r)} \\[1em]
    &+ \kappa_d^2\left(\dfrac{\lambda R M}{2\sqrt{d}}\right)^d M \lambda \varepsilon \sqrt{d} \\[1em]
    &+ \kappa_d^2\left(\dfrac{\lambda RM}{2\sqrt{d}}\right)^d \left(\sum_{n=0}^{d-1} \dfrac{1}{n!}\left[ \dfrac{\delta}{4 d^{3/2}} \right]^n\right)^Me^{-M \delta/4d^{3/2}} \\[1em]
    &+ 2 \varepsilon^{-2d} \left(\sum_{n=0}^{2d}\binom{2d}{n}(4R)^n \right)e^{-M\delta^2/2}.
\end{align*}
The above expression is minimized at
\begin{equation*}
    \varepsilon_0 = \left(2\kappa_d^{-2} Re^{-M \delta^2/2}\left(\dfrac{2\sqrt{d}}{\lambda R M}\right)^{d + 1}\sum_{n=0}^{2d}\binom{2d}{n}(4R)^n\right)^{1/(2d + 1)},
\end{equation*}
and so we have that
\begin{align*}
    \mathbb{P}\left[\sup_{x, x' \in \mathcal{X}} \left|k_M(x, x') - k_\infty(x, x')\right| > \delta\right] &< c_1M^{d-1}e^{-2 \lambda M(R - r)} + c_2M^{d + d/(2d + 1)}e^{-M\delta^2 /(4d + 2)} \\
    &+ c_3 M^{d} e^{M (\alpha(\delta) - \delta/4d^{3/2})} + c_4 M^{d + d/(2d + 1)} e^{- M\delta^2/(4d + 2)}
\end{align*}
for constants $c_1, c_2, c_3, c_4$ not depending on $M, \delta$ and function
\begin{equation*}
    \alpha(\delta) = \log\left(\sum_{n=0}^{d-1}\dfrac{1}{n!} \left[\dfrac{\delta}{4d^{3/2}}\right]^n\right)
\end{equation*}
which satisfies $\alpha(\delta) - \delta/4d^{3/2} < 0$ and $(\alpha(\delta) - \delta/4d^{3/2}) \to 0$ slower than $-M\delta^2/(4d+2) \to 0$ as $\delta \to 0$. Thus, letting $R - r > 0$ be sufficiently small, it holds for small enough $\delta > 0$ that
\begin{equation*}
    \mathbb{P}\left[\sup_{x, x' \in \mathcal{X}}\left|k_M(x, x') - k_\infty(x, x')\right| > \delta\right] \in \mathcal{O}\left(M^{d + d/(2d + 1)}e^{- M\delta^2/(4d + 2)}\right),
\end{equation*}
completing our proof of the rate of uniform convergence of the uniformly rotated Mondrian kernel to its limiting kernel. \hfill $\square$

\section{EXPERIMENTS CODE\label{code}}

The following \href{https://github.com/COsborne25/uniformly-rotated-mondrian-kernel}{link} is to a GitHub repository with the code to replicate our experiments. The \texttt{README.md} file contains full information about the required packages and their respective versions at the time of figure generation.

\vfill

\end{document}